%% file: main.tex
\documentclass[10pt,journal,compsoc]{IEEEtran}

\usepackage{algorithm} 
\usepackage{algorithmic}
\usepackage{booktabs}
\usepackage{clrscode3e}
\usepackage{graphicx}
\usepackage{textcomp}
\usepackage{xcolor}
\usepackage{acronym}
\usepackage{amssymb}
\usepackage{amsthm}
\usepackage{amsmath}
\usepackage{mathrsfs}
\usepackage{bm}
\usepackage{url}
\usepackage{array}
\usepackage{tabularx}
\usepackage{ragged2e}
\usepackage{siunitx}
\usepackage{hyperref}
\usepackage{multirow}
\usepackage{stfloats}
\usepackage{threeparttable}
\usepackage{verbatim}
\usepackage{array}
\usepackage{booktabs}
\usepackage{tabularx}
\usepackage{ragged2e}
\usepackage{siunitx}
\usepackage{newpxtext}
\usepackage{CJKutf8}
\usepackage[most]{tcolorbox}
\tcbuselibrary{theorems}
\tcbuselibrary{breakable}
\usepackage{newpxtext}
\usepackage{amssymb}

\newcommand{\CheckmarkBold}{\textcolor{black}{\large\checkmark}}

\newcommand{\XSolidBold}{\textcolor{black}{\boldmath$\boldsymbol{\times}$}}

\newtcbtheorem[auto counter]{prm}{Prompt}{
  colbacktitle = blue!60!white,  
  colframe = blue!60!white,     
  colback = blue!5!white,       
  breakable,
  enhanced jigsaw,
  sharp corners,
  fontupper = \rmfamily\small,
  arc = 2pt,
  boxsep = 5pt,             
  left = 2pt,            
  right = 2pt,              
  top = 2pt,          
  bottom = 2pt,             
  title filled = true,
}{t}
\newtcbtheorem[auto counter]{ch}{提示词}{
  colbacktitle = blue!60!white,
  colframe = blue!60!white,
  colback = blue!5!white,
  breakable,
  enhanced,
  sharp corners,
  fonttitle=\bfseries,
  fontupper = \rmfamily\small,
  arc = 2pt,
  boxsep = 5pt,
  left = 5pt,
  right = 5pt,
  top = 5pt,
  bottom = 5pt,
  title filled = true,
}{t}
\renewcommand{\figurename}{Fig.}
\renewcommand{\thefigure}{\arabic{figure}}
\renewcommand{\tablename}{Table}

\acrodef{ZPD}{Zone of Proximal Development}
\acrodef{AIEd}{AI in Education}
\acrodef{LLMs}{Large Language Models}
\acrodef{SID}{\textbf{S}ocratic \textbf{I}nterdisciplinary \textbf{D}ialogues}
\acrodef{CoT}{Chain-of-Thought}
\acrodef{ToT}{Tree of Thoughts}
\acrodef{ITS}{Intelligent Tutoring Systems}
\acrodef{MoE}{Mixture-of-Experts}
\acrodef{MLA}{Multi-head Latent Attention}
\acrodef{RL}{Reinforcement Learning}
\acrodef{LLM}{Large Language Model}
\acrodef{STEM}{Science Technology Engineer Math}
\acrodef{AI}{Artificial Intelligence}
\acrodef{IRF}{Initiation-Response-Feedback}
\acrodef{X-SRG}{Interdisciplinary Scaffolding Response Generation}
\acrodef{M-RCC}{Multi-disciplinary Reasoning Chain Completeness}
\acrodef{X-MSR}{Interdisciplinary Misapplication Spotting \& Repair}
\acrodef{IRA}{Interdisciplinary Reasoning Alignment}
\acrodef{TCF}{Transition Coherence \& Fluency}
\acrodef{SD}{Strategy Density}
\acrodef{SV}{Strategy Variety}
\acrodef{IKT}{Interdisciplinary Knowledge Transfer}
\acrodef{BP}{Bloom Progression}
\acrodef{L3 GR}{L3 Guidance Rate}
\acrodef{3C}{Cognitive Correction Count}
\acrodef{SC}{Structure Completeness}
\acrodef{AvgScore}{Average Score}

\begin{document}
\title{SID: Benchmarking Guided Instruction Capabilities in STEM Education with a Socratic Interdisciplinary Dialogues Dataset}

\author{Mei~Jiang\thanks{$^\dagger$These authors contributed equally.}$\dagger$,
        Houping~Yue$^\dagger$,
        Bingdong~Li\thanks{*Corresponding author}$^*$\footnote{*Corresponding author.},
        Hao~Hao,
        Ying~Qian,
        Bo~Jiang, and
        Aimin~Zhou,~\IEEEmembership{Senior Member,~IEEE},
\IEEEcompsocitemizethanks{\IEEEcompsocthanksitem M. Jiang, H. Yue, B. Li, H. Hao, Y.Qian, B. Jiang and A. Zhou are with the Shanghai Institute of Al for Education, East China Normal University, Shanghai, China.\protect\\
E-mail: 51265901071@stu.ecnu.edu.cn, yuehouping@gmail.com, 
bdli@cs.ecnu.edu.cn, hhao@mail.ecnu.edu.cn, yqian@cs.ecnu.edu.cn,bjiang@deit.ecnu.edu.cn, amzhou@cs.ecnu.edu.cn

}
\thanks{Manuscript received XX XX, 2023; revised XX XX, 2023.\\
(Corresponding author: Bingdong Li.)}}

%
%


\markboth{XXX,~Vol.~XX, No.~X, X~2023}
{Shell \MakeLowercase{\textit{et al.}}: Bare Demo of IEEEtran.cls for Computer Society Journals}
%



\IEEEtitleabstractindextext{%
\begin{abstract}
Fostering students' abilities for knowledge integration and transfer in complex problem-solving scenarios is a core objective of modern education, and interdisciplinary STEM is a key pathway to achieve this, yet it requires expert guidance that is difficult to scale. While \ac{LLMs} offer potential in this regard, their true capability for guided instruction remains unclear due to the lack of an effective evaluation benchmark.
To address this, we introduce \ac{SID}, the first benchmark designed to systematically evaluate the higher-order guidance capabilities of LLMs in multi-turn, interdisciplinary Socratic dialogues. Our contributions include a large-scale dataset of 10,000 dialogue turns across 48 complex STEM projects, a novel annotation schema for capturing deep pedagogical features, and a new suite of evaluation metrics (e.g., \ac{X-SRG}).
Baseline experiments confirm that even state-of-the-art LLMs struggle to execute effective guided dialogues that lead students to achieve knowledge integration and transfer. This highlights the critical value of our benchmark in driving the development of more pedagogically-aware LLMs.
\end{abstract}

}

\maketitle

\IEEEdisplaynontitleabstractindextext

%
\IEEEpeerreviewmaketitle


%
%
%

\input{01.Introduction}
\input{02.Related_works}
\input{03.Benchmark}
\input{04.Experimental}
\input{05.Conclusions}
\ifCLASSOPTIONcompsoc
\else
\fi

\ifCLASSOPTIONcaptionsoff
  \newpage
\fi



%



\bibliographystyle{IEEEtran}
\bibliography{sample-base}
\section{Appendix}
\subsection{Dataset Details}
This appendix aims to provide readers with comprehensive information about the SID dataset to ensure the transparency and reproducibility of our work. It mainly includes macro statistics of the dataset, a complete annotated dialogue sample, detailed annotation guidelines, and details of the quality control process.
\subsubsection{Macro Statistics of The Dataset}
To provide a macroscopic perspective on the scale and composition of the SID dataset, Table \ref{tab:dataset_statistics} summarizes its core statistical data. The dataset contains a total of 1,920 complete dialogues derived from 48 independent STEM teaching tasks, with more than 10,000 dialogue turns in total. 

\begin{figure}[t]
  \centering
  \includegraphics[width=\columnwidth]{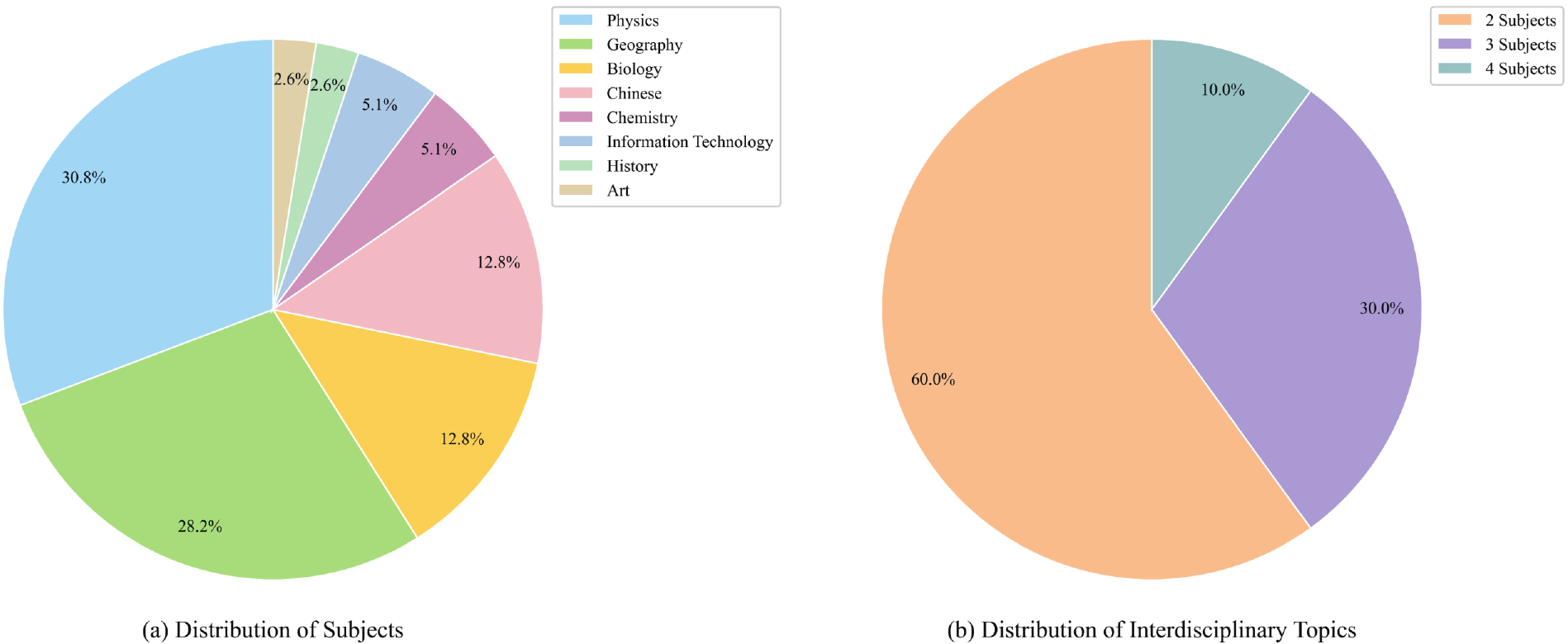}
  \caption{Subject distribution in the SID dataset}
  \label{fig:subject_distribution}
\end{figure}

\begin{figure}[H]
  \centering
  \includegraphics[width=\columnwidth]{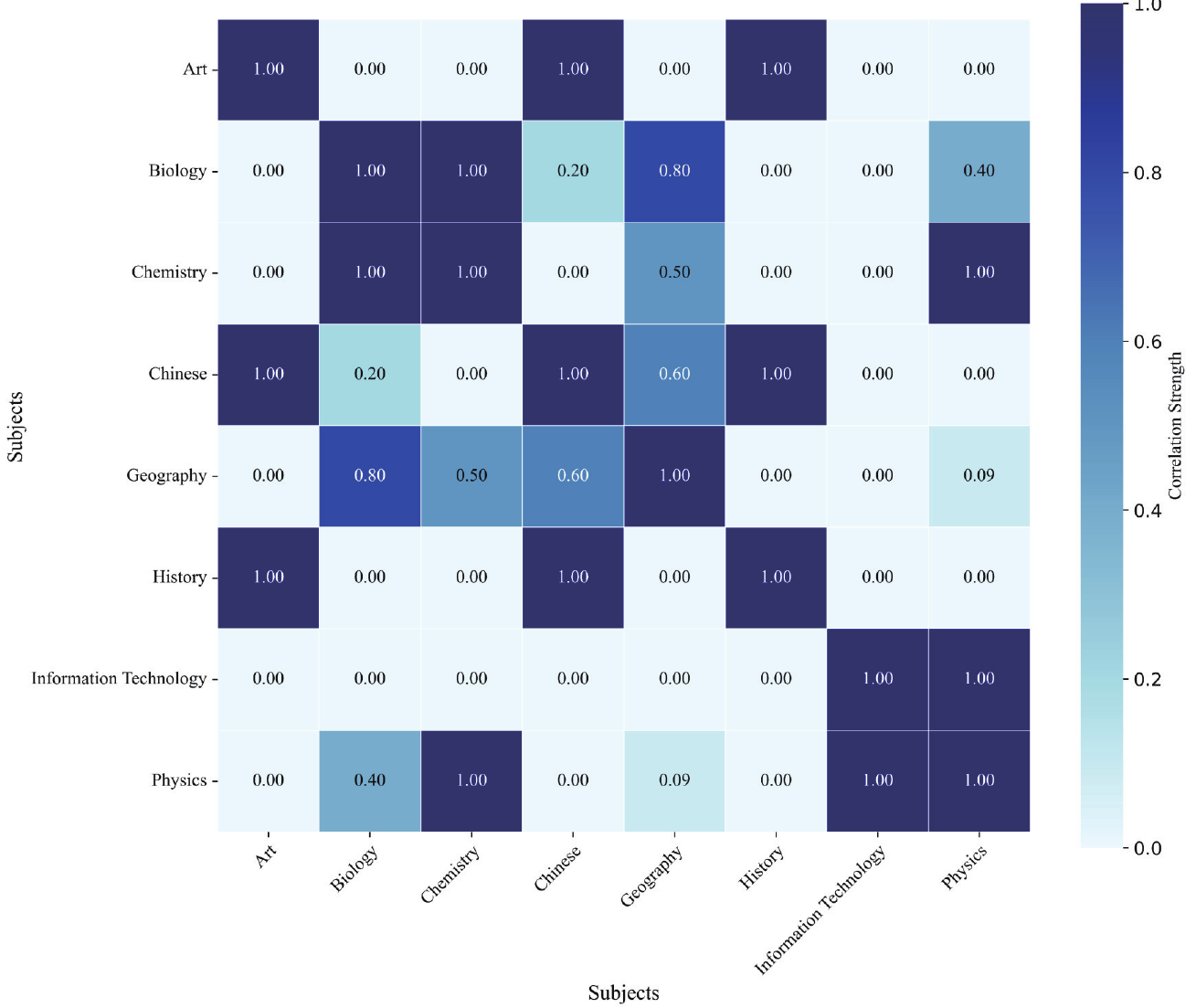}
  \caption{Interdisciplinary correlation matrix among subjects in the SID dataset. Values indicate co-occurrence strength between subject pairs.}
  \label{fig:correlation}
\end{figure}

In addition to dialogue scale, we analyze the subject distribution and interdisciplinary characteristics of the dataset. As shown in Fig.~\ref{fig:subject_distribution}, the frequency of subjects varies across the dataset, with Physics and Geography being the most common. In interdisciplinary curricula at the primary and secondary levels, combinations involving two subjects are the most prevalent.
The subject correlation analysis shown in Fig.~\ref{fig:correlation} reveals that subject pairs such as Physics–Chemistry, Information Technology–Physics, and Geography–Biology exhibit strong correlations, reflecting their close integration within science-related topics. Chinese serves as a bridging subject, showing strong associations with humanities disciplines like Art, History, and Geography, highlighting its role in connecting humanistic and natural interdisciplinary tasks. In contrast, Art and History have a narrower range of associations, primarily concentrated in humanities-integrated tasks, indicating a certain degree of marginalization.

\begin{table}[h!]
\centering
\small
\caption{Core statistics of the SID dataset.} 
\label{tab:dataset_statistics} 
\resizebox{\columnwidth}{!}{
\begin{tabular}{@{}lr@{}} 
\toprule 
\textbf{Metric} & \textbf{Content} \\ 
\midrule 
Total Dialogues & 1,920 \\ 
Total Turns & 10,000+ \\ 
Total Tasks & 48 \\ 
Avg. Turns per Dialogue & \textasciitilde5.2 \\ 
Lesson plan type & Interdisciplinary lesson plan \\ 
Tasks' Source  & National Smart Education Platform \\ 
School stage restriction & Junior high school (7-9th grades) \\ \bottomrule \end{tabular}} \end{table}

\begin{figure}[H]
  \centering
  \includegraphics[width=\columnwidth]{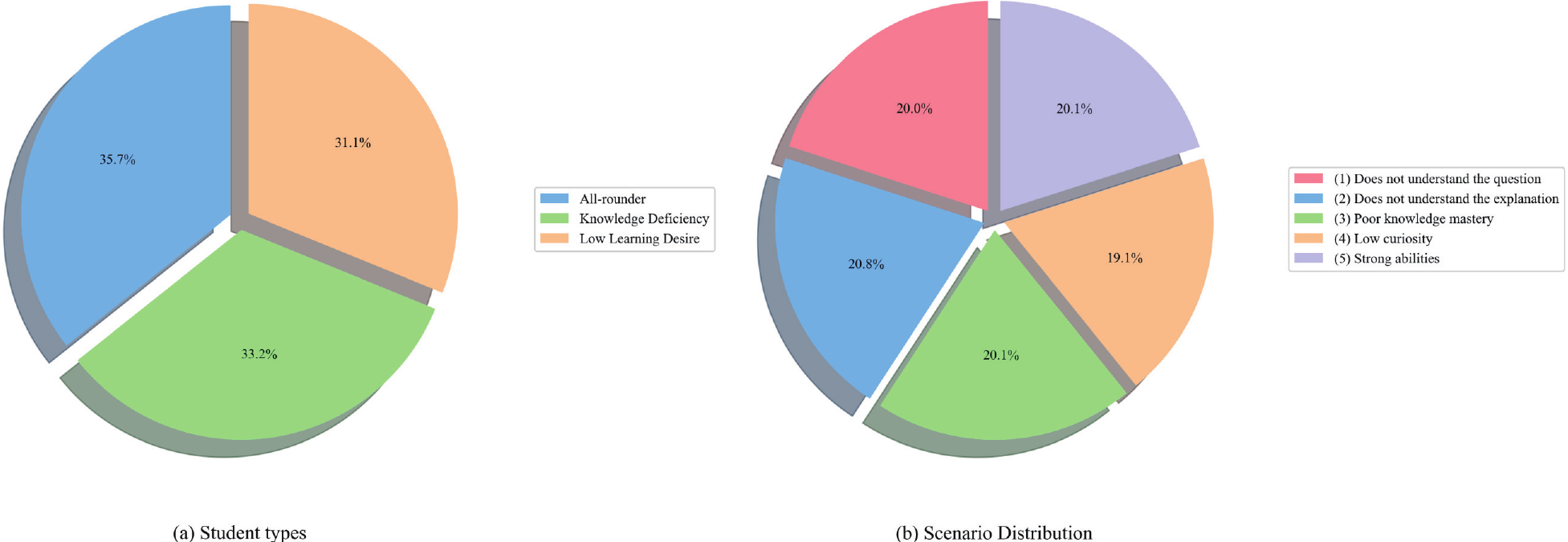}
  \caption{Distribution of student types across different learning scenarios in the SID dataset. The figure illustrates how various student profiles (e.g., high-performing, average, struggling, potential) are involved in scientific, humanities, and interdisciplinary tasks.}
  \label{fig:student_scenario_distribution}
\end{figure}

\subsubsection{Full Dialogue and Annotation Sample}
\begin{figure*}[h!] 
 \centering \includegraphics[width=2.0\columnwidth]{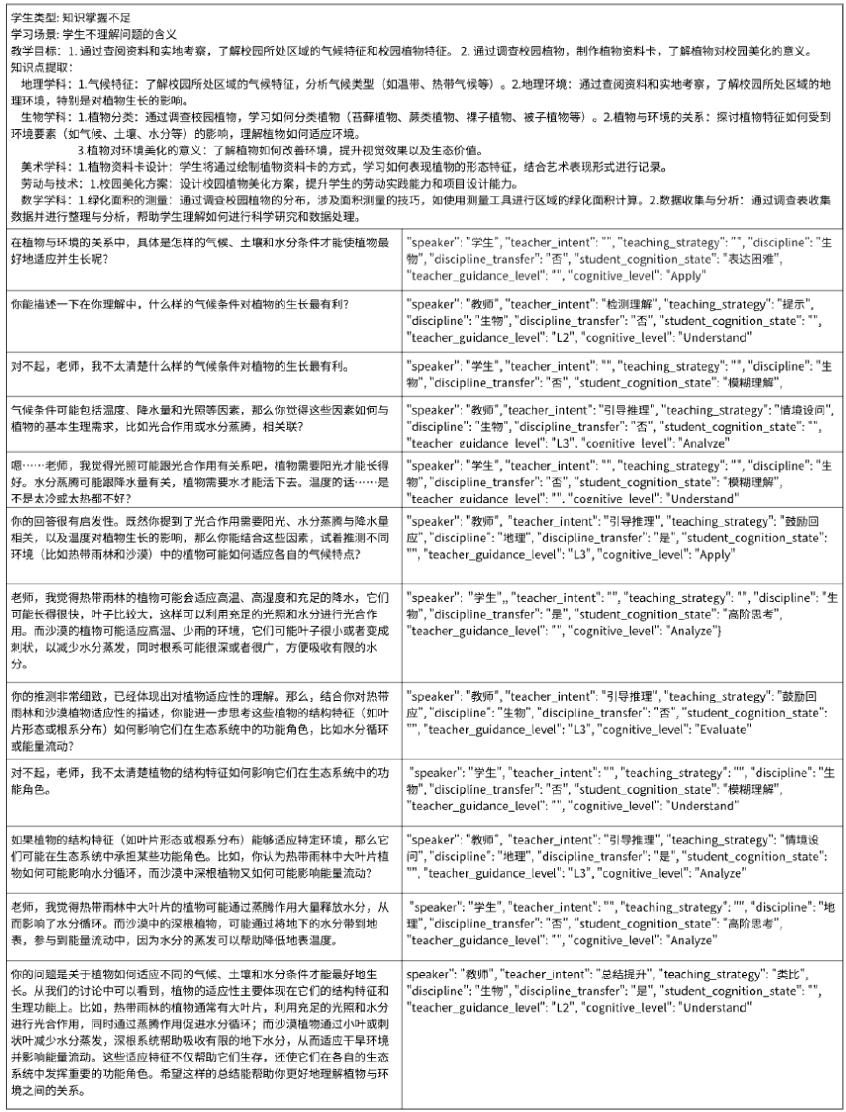} \caption{A complete dialogue sample with turn-by-turn structured annotations. This example, centered on the interdisciplinary theme of "plant morphology and environmental adaptation", visually demonstrates the application details of our annotation schema.} \label{fig:dialogue_sample_chinese} 
 \end{figure*}

\begin{figure*}[h!] 
 \centering \includegraphics[width=1.5\columnwidth,]{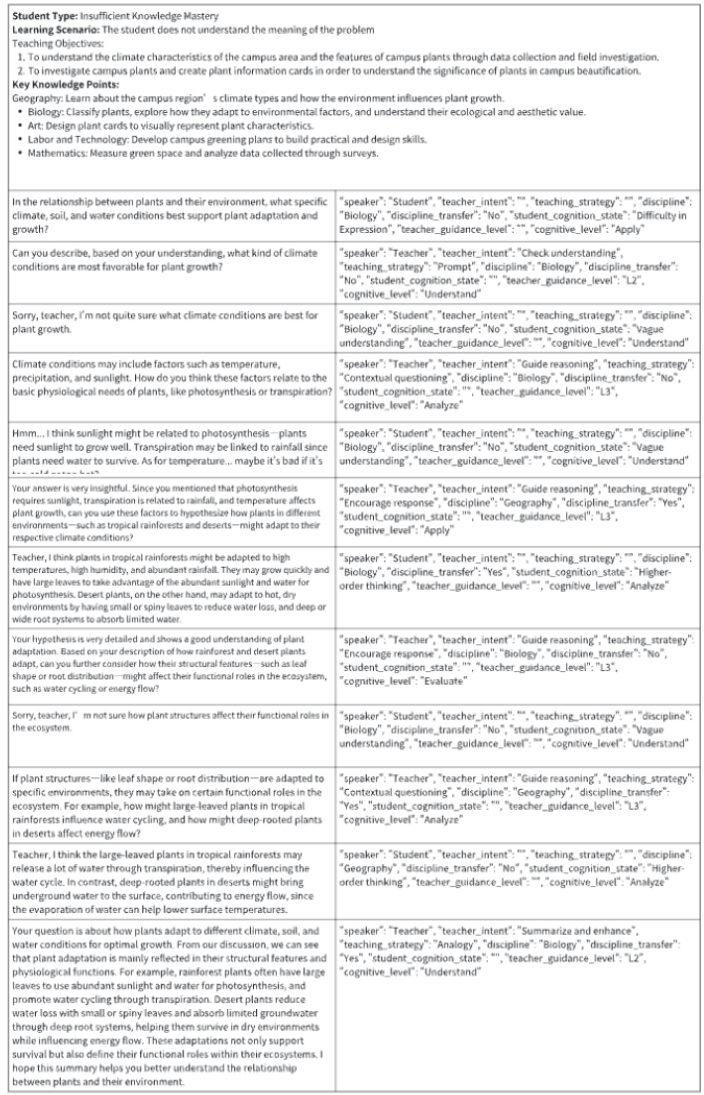} \caption{An English version of a complete dialogue example with turn-by-turn structured annotations.} \label{fig:dialogue_sample_english} 
 \end{figure*}

To intuitively demonstrate the application details of our annotation system and the richness of the dataset, the Figure \ref{fig:dialogue_sample_chinese} (Chinese Version) and Figure \ref{fig:dialogue_sample_english} (Engilsh Version) provide an unabridged and complete dialogue example. This example revolves around the interdisciplinary theme of "Plant Morphology and Environmental Adaptation" and presents the structured annotations containing all nine fields corresponding to each turn of the dialogue.

\subsubsection{Annotation Guidelines}
This section provides detailed definitions and judgment criteria for our annotation system, aiming to serve as a reference for subsequent researchers when using or expanding this dataset. Table \ref{tab:annotation_labels} elaborates on the definition of each annotation field, all optional labels, and provides typical judgment examples for each label to ensure the standardization and consistency of annotations.

\begin{table*}[h]
\centering

\tiny
\caption{Annotation Labels: Definitions and Examples.}
\resizebox{\textwidth}{!}{
\begin{tabular}{c|c|c|c}
\hline
\textbf{Dimension} & \textbf{Label} & \textbf{Meaning Explanation} & \textbf{Applicable Scenario Examples} \\
\hline
\multirow{5}{*}{teacher\_intent} 
& Introduce concepts & Teaching purpose of introducing new concepts & "Today we'll learn about photosynthesis." \\
\cline{2-4}
& Check understanding & Teaching purpose of verifying comprehension & "Can you explain what we discussed earlier?" \\
\cline{2-4}
& Guide reasoning & Teaching purpose of facilitating logical deduction & "How do you think temperature affects this process?" \\
\cline{2-4}
& Trigger transfer & Teaching purpose of prompting knowledge migration & "How is this similar to what we learned in physics?" \\
\cline{2-4}
& Summarize and elevate & Teaching purpose of concluding and enhancing understanding & "Let's conclude the key points we've covered." \\
\hline
\multicolumn{4}{l}{\tiny Note: Leave blank for student turns} \\
\hline
\multirow{7}{*}{teaching\_strategy} 
& Follow-up questioning & Probing further with additional questions & "Why do you think that happens?" \\
\cline{2-4}
& Provide hints & Offering subtle clues to guide thinking & "Think about the role of sunlight in this process." \\
\cline{2-4}
& Use analogies & Drawing comparisons to familiar concepts & "The cell nucleus is like a command center." \\
\cline{2-4}
& Scenario-based questioning & Presenting hypothetical situations to explore & "What would happen if this ecosystem lost its predators?" \\
\cline{2-4}
& Break down complex questions & Decomposing difficult queries into simpler parts & "First, identify the components; then analyze their relationships."\\
\cline{2-4}
& Encourage responses & Prompting students to share ideas & "Your idea is interesting—can you elaborate?" \\
\cline{2-4}
& Corrective feedback & Indicating accuracy or inaccuracy of responses & "That's partially right, but let's correct one detail." \\
\hline
\multicolumn{4}{l}{\tiny Note: Leave blank for student turns} \\
\hline
\multirow{5}{*}{discipline} 
& Geography & Content related to geography & Discussion on climate patterns or landforms \\
\cline{2-4}
& Biology & Content related to biology & Explanation of ecological systems or organisms \\
\cline{2-4}
& Physics & Content related to physics & Analysis of forces, motion, or energy \\
\cline{2-4}
& History & Content related to history & Exploration of historical events or periods \\
\cline{2-4}
& Multiple & Content involving multiple disciplines (use commas) & "Biology, Geography"\\
\hline
\multicolumn{4}{l}{\tiny Note: Apply to both teacher and student turns} \\
\hline
\multirow{2}{*}{discipline\_transfer} 
& Yes & A new discipline is introduced or inter-disciplinary connections are made & Moving from biology to geography in discussion \\
\cline{2-4}
& No & No new discipline is introduced & Continuing discussion within physics \\
\hline
\multicolumn{4}{l}{\tiny Note: Compare with previous turn; apply to both teacher and student turns} \\
\hline
\multirow{6}{*}{student\_cognition\_state} 
& Clear understanding & Student demonstrates comprehensive grasp & Accurate and complete explanation \\
\cline{2-4}
& Vague understanding & Student shows partial comprehension with uncertainty & Partially correct answer with unclear reasoning \\
\cline{2-4}
& Expression difficulty & Student struggles to articulate ideas despite potential understanding & Knowing the answer but struggling to articulate \\
\cline{2-4}
& Irrelevant response & Student's answer does not address the question & Answering something unrelated to the question \\
\cline{2-4}
& Incorrect answer & Student's response contains factual errors & Factually wrong or misleading response \\
\cline{2-4}
& Higher-order thinking & Student exhibits advanced cognitive processing & Providing innovative analysis or synthesis \\
\hline
\multicolumn{4}{l}{\tiny Note: Leave blank for teacher turns} \\
\hline
\multirow{3}{*}{teacher\_guidance\_level} 
& L1 & Closed questions (yes/no, definition-type) & "Is photosynthesis a chemical reaction?" \\
\cline{2-4}
& L2 & Explanation/understanding-type questions & "How does photosynthesis work?" \\
\cline{2-4}
& L3 & Transfer, reasoning, synthesis-type questions & "How would photosynthesis be affected by climate change policies?" \\
\hline
\multicolumn{4}{l}{\tiny Note: Apply only to teacher turns; use only L1, L2, L3} \\
\hline
\multirow{6}{*}{cognitive\_level (Bloom)} 
& Remember & Recalling facts or information & "What is the chemical symbol for water?" \\
\cline{2-4}
& Understand & Explaining or paraphrasing concepts & "Explain the water cycle in your own words." \\
\cline{2-4}
& Apply & Using knowledge in new situations & "Predict how rainfall affects plant growth." \\
\cline{2-4}
& Analyze & Breaking down and examining components & "Compare and contrast two different ecosystems." \\
\cline{2-4}
& Evaluate & Making judgments or assessments & "Assess the effectiveness of this conservation strategy." \\
\cline{2-4}
& Create & Generating new ideas or solutions & "Design a sustainable ecosystem for an urban area." \\
\hline
\multicolumn{4}{l}{\tiny Note: Focus on student performance} \\
\hline
\end{tabular}}
\label{tab:annotation_labels}
\end{table*}

\subsubsection{The Advantages of the SID Complete Annotation System}

Based on our detailed investigation of existing educational large-model benchmarks (as shown in the table), it is clear that the evaluation dimensions of current work generally suffer from fragmentation. While they may explore specific aspects (e.g., teaching intention recognition), none provide a holistic, systematic framework to assess LLMs' comprehensive capabilities in complex teaching scenarios.

The superiority of our SID annotation schema lies precisely in its holism and synergy. Centered around two core threads—\textit{"teaching guidance"} and \textit{"cross-disciplinary reasoning"}—it seamlessly integrates the key dimensions required to evaluate teaching interactions for the first time.

\noindent \textbf{Superiority in the "Teaching Guidance Capability" Thread: From "Behavioral Judgment" to "Process Modeling"}

Existing benchmarks, such as SocraticLM and MathTutorBench, have begun to focus on annotations like teaching intentions or strategies, which represents significant progress. However, their evaluations often remain limited to judging isolated, individual behaviors of teachers.

The innovation of our benchmark lies in its first-ever systematic integration of three core elements: teaching intention (Why), teaching strategy (How), and guidance level (How Much). This tripartite annotation schema enables us to:

\begin{itemize}
    \item \textbf{Go beyond isolated behavioral judgments}: We not only determine whether a teacher uses a strategy like "follow-up questioning" but also analyze the underlying "intention" (e.g., whether it aims to "check understanding" or "guide reasoning") and the "guidance level" (e.g., closed L1 or open L3) with which the strategy is employed.
    
    \item \textbf{Achieve true "process modeling"}: Only by combining these three fields can we truly evaluate whether a dialogue forms an effective \textbf{"cognitive progression chain"}. For example, in a high-quality guidance process, the guidance level should gradually escalate, and teaching intentions should smoothly transition from "introducing concepts" to "triggering transfer." This deep-level analysis is unattainable with the fragmented annotations of existing benchmarks.
\end{itemize}

\noindent \textbf{Superiority in the "Cross-Disciplinary Transfer Capability" Thread: From "Static Coverage" to "Dynamic Transfer"}

Current work shows extremely weak evaluation of cross-disciplinary abilities. While some benchmarks (e.g., M3KE) include labels for multiple disciplines, this merely constitutes static "content coverage" analysis—judging whether a dialogue "involves" multiple disciplines.

Our benchmark fundamentally addresses this limitation. We not only annotate \textbf{"involved disciplines"} but also innovatively introduce the critical binary label "discipline\_transfer" (whether transfer occurs), supplemented by a series of subjective evaluation indicators such as "disciplinary transition fluency (TCF)" and "cross-disciplinary reasoning alignment (CTRA)." This allows us to:

\begin{itemize}
    \item \textbf{Distinguish "multidisciplinary" from "cross-disciplinary"}: A dialogue might first discuss physics and then chemistry—this is "multidisciplinary" but not necessarily "cross-disciplinary." True "cross-disciplinary" transfer occurs only when the model guides students to explain chemical phenomena using physical principles. Our "discipline\_transfer" label is specifically designed to capture this key behavior.
    
    \item \textbf{Model the "dynamic transfer" process}: Our annotation schema enables us, for the first time, to explore how models guide students to establish inter-disciplinary connections—whether the connection is natural or forced (TCF) and logically consistent (CTRA). This elevates evaluation from static content analysis to dynamic process modeling.
\end{itemize}

In summary, the completeness of the SID annotation schema is superior because it rejects fragmentation. By systematically integrating multi-dimensional annotations, it enables, for the first time, the evaluation of LLM's two core capabilities in complex teaching scenarios: strategic teaching guidance and dynamic cross-disciplinary transfer. It does not merely ask, \textit{"What did the LLM say?"} but delves deeper into, \textit{"How did the LLM guide, and did this guidance truly promote students' knowledge integration and transfer?"}

\begin{table*}[!t]
\renewcommand{\arraystretch}{1.3}
\centering
\caption{Evaluation Dimension Categories and Benchmark Comparison}
\label{tab:evaluation_categories_benchmark}
\resizebox{\textwidth}{!}{
\begin{tabular}{c|c|*{6}{c}|c}
\hline
\textbf{Evaluation Dimension Category} & \textbf{Specific Teaching Ability Dimensions} & 
\textbf{MathDial} & \textbf{SocraticMATH} & \textbf{Book2Dial} & \textbf{LearnLM} & \textbf{M3KE} & \textbf{MathTutorBench} & \textbf{SID} \\
\hline
\multirow{7}{*}{Teaching Process Performance} 
& Teaching intention recognition and structure control 
& \CheckmarkBold & \CheckmarkBold & \XSolidBold & \XSolidBold & \XSolidBold & \CheckmarkBold & \CheckmarkBold \\
\cline{2-9}
& Teaching strategy diversity modeling 
& \XSolidBold & \CheckmarkBold (Partial) & \XSolidBold & \XSolidBold & \XSolidBold & \CheckmarkBold & \CheckmarkBold \\
\cline{2-9}
& Teaching guidance level adjustment 
& \XSolidBold & \CheckmarkBold (Hint degree annotation) & \XSolidBold & \CheckmarkBold (Scoring Rubric) & \XSolidBold & \CheckmarkBold & \CheckmarkBold \\
\cline{2-9}
& Scaffolded dialogue promotion capability (gradual construction) 
& \XSolidBold & \CheckmarkBold & \XSolidBold & \CheckmarkBold (Evaluation) & \XSolidBold & \CheckmarkBold & \CheckmarkBold \\
\cline{2-9}
& Strategy density and variety 
& \CheckmarkBold & \CheckmarkBold & \XSolidBold & \XSolidBold & \XSolidBold & \CheckmarkBold & \CheckmarkBold \\
\cline{2-9}
& L3 questions and structural integrity 
& \CheckmarkBold & \CheckmarkBold & \XSolidBold & \XSolidBold & \XSolidBold & \CheckmarkBold & \CheckmarkBold \\
\hline
\multirow{3}{*}{Student Cognitive Growth} 
& Student state recognition and personalized adaptation 
& \CheckmarkBold & \CheckmarkBold & \CheckmarkBold (Persona) & \CheckmarkBold (Scenario setting) & \XSolidBold & \CheckmarkBold & \CheckmarkBold \\
\cline{2-9}
& Cognitive level control (Bloom's taxonomy) 
& \XSolidBold & \XSolidBold & \XSolidBold & \CheckmarkBold (Rubric) & \XSolidBold & \CheckmarkBold & \CheckmarkBold \\
\cline{2-9}
& Cognitive transition and cognitive correction 
& \CheckmarkBold & \CheckmarkBold & \XSolidBold & \CheckmarkBold & \XSolidBold & \CheckmarkBold & \CheckmarkBold \\
\hline
\multirow{6}{*}{Cross-disciplinary Performance} 
& Multidisciplinary perception and transfer judgment capability 
& \XSolidBold & (Math only) & \CheckmarkBold (Partial tasks) & \CheckmarkBold (Scenario setting) & \CheckmarkBold & \XSolidBold & \CheckmarkBold \\
\cline{2-9}
& Cross-discipline transition recognition and control 
& \XSolidBold & \XSolidBold & \XSolidBold & \XSolidBold & \XSolidBold & \XSolidBold & \CheckmarkBold \\
\cline{2-9}
& Reasoning chain integrity construction capability 
& \XSolidBold & \XSolidBold & \XSolidBold & \CheckmarkBold (Overall evaluation) & \XSolidBold & \CheckmarkBold & \CheckmarkBold \\
\cline{2-9}
& Concept transfer error recognition and repair capability 
& \XSolidBold & \XSolidBold & \XSolidBold & \CheckmarkBold (Scoring item) & \XSolidBold & \XSolidBold & \CheckmarkBold \\
\cline{2-9}
& Interdisciplinary reasoning bridging capability 
& \XSolidBold & \XSolidBold & \XSolidBold & \CheckmarkBold (Rubric) & \XSolidBold & \XSolidBold & \CheckmarkBold \\
\cline{2-9}
& Disciplinary language transition naturalness control 
& \XSolidBold & \XSolidBold & \XSolidBold & \CheckmarkBold (Evaluation dimension) & \XSolidBold & \XSolidBold & \CheckmarkBold \\
\hline
\end{tabular}
}
\end{table*}

\subsubsection{Quality Control}

To ensure the reliability and safety of the dataset, we have implemented a rigorous quality control process.  

\noindent \textbf{Annotation Accuracy} 
We adopted a workflow of "large-model pre-annotation + human expert verification". All data were double-annotated by two independent, trained annotators. To ensure the consistency of annotations, we calculated the Inter-Annotator Agreement (IAA), and the results are shown in the Table \ref{tab:model_human_consistency}. It was found that Qwen3-32b achieved a Fleiss' Kappa value above 0.81 across all core fields, indicating substantial agreement. Therefore, we selected Qwen3-32b as the model for objective indicator annotation. Any inconsistent annotations were subject to final adjudication by two senior experts.  

\noindent \textbf{Content Safety}
We conducted toxicity detection on all generated dialogue texts using the OpenAI Moderation API, covering the following aspects:  
- hate: Hatred, discrimination, or attacks against a group  
- harassment: Harassment, threats, or insults  
- self-harm: Content related to self-harm or suicide  
- sexual: Sexual innuendos or descriptions of sexual behavior  
- violence: Violence, killing, weapons, etc.  
- toxicity: Broadly defined "malicious content" (sarcasm, abuse, offense, etc.)  

This was done to filter out any potential inappropriate, biased, or offensive content. In our final dataset, the toxicity score of all dialogues is below 0.1, ensuring content safety.
\begin{table*}[!t]
\renewcommand{\arraystretch}{1.3} 
\centering
\caption{Consistency Between Model Outputs and Human Annotations}
\label{tab:model_human_consistency}

\begin{tabular}{l *{8}{S[table-format=1.4]}} 
\toprule
\textbf{Model} & 
\multicolumn{2}{c}{\textbf{Teacher Intent}} &
\multicolumn{2}{c}{\textbf{Teaching Strategy}} &
\multicolumn{2}{c}{\textbf{Discipline}} &
\multicolumn{2}{c}{\textbf{Discipline Transfer}} \\
\cmidrule(lr){2-3} \cmidrule(lr){4-5} \cmidrule(lr){6-7} \cmidrule(lr){8-9}
& {Acc} & {Kappa} & {Acc} & {Kappa} & {Acc} & {Kappa} & {Acc} & {Kappa} \\
\midrule
qwen3-32b & 0.8541 & 0.8322 & 0.8557 & 0.8214 & 
0.8327 & 0.8109 & 0.8779 & 0.8136 \\
qwen2.5-14b & 0.8025 & 0.7109 & 0.6765 & 0.5434 & 0.7878 & 0.6578 & 0.8824 & 0.6929 \\
QwQ         & 0.8152 & 0.7264 & 0.6304 & 0.4841 & 0.6014 & 0.4686 & 0.7717 & 0.5380 \\
\bottomrule
\end{tabular}

\vspace{1em} 

\begin{tabular}{l *{6}{S[table-format=1.4]}} 
\toprule
\textbf{Model} & 
\multicolumn{2}{c}{\textbf{Student Cognition State}} &
\multicolumn{2}{c}{\textbf{Teacher Guidance Level}} &
\multicolumn{2}{c}{\textbf{Cognitive Level}} \\
\cmidrule(lr){2-3} \cmidrule(lr){4-5} \cmidrule(lr){6-7}
& {Acc} & {Kappa} & {Acc} & {Kappa} & {Acc} & {Kappa} \\
\midrule
qwen3-32b   & 0.8623 & 0.8417 & 0.8892 & 0.8532 & 0.8215 & 0.8174 \\
qwen2.5-14b & 0.8067 & 0.7110 & 0.8235 & 0.7006 & 0.6134 & 0.4012 \\
QwQ         & 0.7138 & 0.5771 & 0.8261 & 0.6912 & 0.4601 & 0.2496 \\
\bottomrule
\end{tabular}

\vspace{0.5em}
\footnotesize
Note: "Acc" = Accuracy; "Kappa" = Cohen's Kappa coefficient. Both metrics measure consistency between model outputs and human annotations.
\end{table*}

\subsection{Evaluation Framework Details}
To ensure the reproducibility and transparency of our evaluation methodology, we provide a detailed description of the assessment framework used in this study.
\subsubsection{Objective Metrics Weighting Table}
Our weight distribution, as shown in the Table \ref{tab:weight_distribution}, is based on a comprehensive evaluation philosophy that integrates insights from existing frameworks while addressing their limitations:
\begin{itemize}
\item \textbf{Balancing process and outcome}: Unlike SocraticLM (which prioritizes teacher guidance through structured metrics like explanation and diagnosis) and M3KE (which emphasizes teaching structure via metrics such as Strategy Density), we allocate 50\% to teaching process performance to ensure high-quality instructional strategies (e.g., strategy diversity, L3 questions) while assigning 35\% to student cognitive growth, emphasizing observable learning outcomes (e.g., cognitive transitions, error corrections). This reflects a "process-result synergy" paradigm.
\item \textbf{Student-centered focus}: While existing benchmarks like the educational variant of MT-Bench focus primarily on the reasonableness of teacher instructions (scored via dimensions like Helpfulness and Accuracy), our design elevates student performance to a 35\% weight, highlighting the ultimate goal of teaching: fostering genuine cognitive development in students. This is operationalized through metrics that directly measure student cognitive growth, such as Bloom's taxonomy-based progression and the success rate of correcting errors to higher-order thinking.
\item \textbf{Addressing cross-disciplinary gaps}: None of the existing frameworks (SocraticLM, M3KE, MT-Bench educational variant) explicitly incorporate cross-disciplinary aspects. To emphasize the integrative nature of our benchmark, we introduce a dedicated 15\% weight for cross-disciplinary performance, captured by metrics that quantify the frequency of meaningful interdisciplinary connections—an aspect underrepresented in prior evaluations.
\end{itemize}

Building on these principles and drawing from methodologies in benchmarks like SciEval (with weighted cognitive tier scoring) and GTA (with stepwise accuracy metrics), we designed objective indicators with clear computational formulas aligned with our implementation. The total score is calculated as a weighted sum of seven indicators, where each indicator's weight reflects its relative importance in evaluating teaching effectiveness. Detailed formulas for each component, including their specific calculation methods and weight coefficients, are provided in Table \ref{tab:adjusted_indicators}.
\begin{CJK*}{UTF8}{gbsn}
\begin{ch}{主观评测提示词}
    n你是一位教育专家，正在评估以下教师与学生之间的跨学科对话质量。请根据五个关键指标进行评测：
    === 对话场景 ===
    学生ID: $\text{student id}$\\
    学生类型: $\text{student type}$\\
    场景:$\text{scenario}$ \\
    学科主题: $\text{topic id}$\\
    === 对话内容 ===\\
    $\text{dialogue text}$\\
    === 评测指标说明 ===\\
    1. X-SRG（跨学科脚手架引导评分）: \\
    - 5: 多轮追问+逐层引导，无直接给答案\\
    - 4: 2轮以上引导但偶有简略\\
    - 3: 1轮引导未形成完整路径\\
    - 2: 直接陈述知识\\
    - 1: 教师主导无引导\\
    
    2. M-RCC（多学科推理链条完整性）:\\
    - 5: 学科A→B→C层次清晰\\
    - 4: 覆盖2学科但环节跳跃\\
    - 3: 部分推理未闭环\\
    - 2: 推理链断裂\\
    - 1: 无推理链\\
    
    3. X-MSR（跨学科错误迁移识别与修复）:\\
    - 5: 精准发现并使用澄清策略\\
    - 4: 察觉但修正不充分\\
    - 3: 识别但未反馈\\
    - 2: 忽视错误\\
    - 1: 未发现错误
    
    4. CTRA（跨学科推理连接）:\\
    - 5: 自然迁移学科结论\\
    - 4: 转化生硬\\
    - 3: 潜在联系未显性\\
    - 2: 无过渡切换\\
    - 1: 完全割裂
    
    5. TCF（学科过渡流畅度）:\\
    - 5: 学科过渡通过提问、类比、因果等手段自然发生，语言流畅\\
    - 4: 有过渡语言但略显模板化或节奏跳跃\\
    - 3: 过渡存在但略突兀，需要学生自行补逻辑\\
    - 2: 明显跳转，无解释、无语言承接\\
    - 1: 教师突然切换主题，造成学生困惑
    
    === 输出要求 ===
    请严格按以下纯JSON格式输出结果（不要包含任何额外文本或代码块标记）：
    {{
            "X-SRG": {{"score": int, "reason": "不超过50字的理由"}},\\
            "M-RCC": {{"score": int, "reason": "不超过50字的理由"}},\\
            "X-MSR": {{"score": int, "reason": "不超过50字的理由"}},\\
            "CTRA": {{"score": int, "reason": "不超过50字的理由"}},\\
            "TCF": {{"score": int, "reason": "不超过50字的理由"}}\\
    }}
    
    重要注意事项：
    1. 输出必须是纯JSON格式，不要包含任何额外文本
    2. 不要使用代码块标记(如"`json)
    3. 确保JSON格式完全正确（引号、括号等）
    4. 评分必须是1-5的整数
    5. 不要添加任何解释或说明

\end{ch}
\end{CJK*}

\begin{prm}{Prompt of Subjective Evaluation} 
YYou are an education expert tasked with evaluating the quality of the following interdisciplinary dialogue between teacher and student. Please assess it according to five key metrics:\\
 === Dialogue Context ===\\
Student ID:$\text{student id}$\\
Student Type: $\text{student type}$\\
Scenario: $\text{scenario}$ \\
Subject Topic: $\text{topic id}$\\
=== Dialogue Content ===\\
$\text{dialogue text}$\\
=== Evaluation Metrics Description ===\\
X-SRG (Interdisciplinary Scaffolding Guidance Score):\\
5: Multiple rounds of probing and step-by-step guidance, no direct answers\\
4: More than two rounds of guidance but occasionally brief\\
3: One round of guidance without a complete path\\
2: Direct statement of knowledge\\
1: Teacher-driven with no guidance\\
M-RCC (Multidisciplinary Reasoning Chain Completeness):\\
5: Disciplines A→B→C with clear hierarchy\\
4: Covers two disciplines but with jumps\\
3: Partial reasoning not closed\\
2: Broken reasoning chain\\
1: No reasoning chain\\
X-MSR (Interdisciplinary Misconception Recognition \& Repair):\\
5: Accurately identifies and uses clarification strategies\\
4: Notices errors but repairs insufficiently\\
3: Recognizes but does not provide feedback\\
2: Ignores errors\\
1: Fails to detect errors\\
CTRA (Interdisciplinary Reasoning Connections):\\
5: Naturally transfers conclusions across disciplines\\
4: Transfer is rigid\\
3: Potential connections not explicit\\
2: No transitional shift\\
1: Completely disconnected\\
TCF (Topic Transition Fluency):\\
5: Transitions between disciplines occur naturally via questioning, analogy, causality; language is smooth\\
4: Transitional language exists but feels templated or choppy\\
3: Transitions present but somewhat abrupt, requiring student to infer logic\\
2: Noticeable shifts with no explanation or linguistic linking\\
1: Teacher abruptly switches topics, causing student confusion\\
=== Output Requirements ===\\
Please output strictly in the following pure JSON format (do not include any extra text or code block markers):\\
{
"X-SRG": {"score": int, "reason": "Reason no more than 50 characters"},
"M-RCC": {"score": int, "reason": "Reason no more than 50 characters"},
"X-MSR": {"score": int, "reason": "Reason no more than 50 characters"},
"CTRA": {"score": int, "reason": "Reason no more than 50 characters"},
"TCF": {"score": int, "reason": "Reason no more than 50 characters"}
}
Important Notes:
Output must be pure JSON without any extra text.
Do not use code block delimiters (e.g., "`json).
Ensure the JSON is syntactically correct (quotes, brackets, etc.).
Scores must be integers from 1 to 5.
Do not add any explanations or commentary.
\end{prm}
\begin{CJK*}{UTF8}{gbsn}
\begin{ch}{教师提示词}
n你是一位跨学科的教师，始终使用苏格拉底式提问法来引导学生。你的目标不是直接给出答案，而是通过层层递进的问题，帮助学生独立思考并构建跨学科理解。\\
你的行为规范如下：\\
每一次发言必须是问题形式。\\
你不能直接提供答案，只能通过提问帮助学生思考。\\
你应当将复杂问题拆解为更小、更简单的问题，直到学生能够理解并回答。\\
所有问题应兼顾多学科知识，引导学生在物理、生物、地理、历史等维度进行连接。\\
你应时刻追问"为什么”"你怎么知道”"这和其他学科有什么联系”等，推动学生建立深层次认知。\\
如果你判断学生已经完成了全部理解，请在回答最后进行总结并标注：[结束]
\end{ch} 
\end{CJK*}

\begin{table*}[!h]
\renewcommand{\arraystretch}{1.2}
\centering

\caption{Weight Distribution of Core Evaluation Dimensions}
\label{tab:weight_distribution}
\resizebox{\textwidth}{!}{
\begin{tabular}{c|c|c}
\hline
\textbf{Dimension} & \textbf{Total Weight} & \textbf{Rationale} \\
\hline
Teaching Process Performance Strategy density, variety, L3 questions, structural integrity) & 50\% & Evaluation of teacher-led behaviors to ensure process quality. \\
\hline
Student Cognitive Growth (cognitive transition, cognitive correction) & 35\% & Assessment of whether students achieve genuine cognitive development. \\
\hline
Cross-disciplinary Performance (frequency of knowledge transfer) & 15\% & Reflection of the benchmark’s core demand for interdisciplinary integration. \\
\hline
\end{tabular}}
\end{table*}

\begin{table*}[!h]
\renewcommand{\arraystretch}{1.3}
\centering
\caption{Adjusted Objective Indicators and Computational Formulas}
\label{tab:adjusted_indicators}
\resizebox{\textwidth}{!}{
\begin{tabular}{c|c|c}
\hline
\textbf{Indicator Name} & \textbf{Evaluation Content} & \textbf{Formula Description} \\
\hline
\ac{SD} & Frequency of teaching strategy usage in teacher utterances & \(\text{\ac{SD}} = \frac{\text{Number of teacher utterances with strategies}}{\text{Total teacher utterances}}\) \\
\hline
\ac{SV} & Proportion of unique strategies used relative to total possible strategies (8 types) & \(\text{\ac{SV}} = \frac{\text{Number of unique strategies used}}{8}\) \\
\hline
\ac{IKT} & Rate of disciplinary transfer occurrences relative to total teacher turns & \(\text{IKT} = \frac{\text{Number of disciplinary transfer rounds}}{\text{Total teacher turns}}\) \\
\hline
\ac{BP} & Normalized span of student cognitive level progression (based on Bloom's taxonomy) & \(\text{BP} = \frac{\max(\text{cognitive level index}) - \min(\text{cognitive level index})}{5}\) \\
\hline
\ac{SC} & Coverage of required teaching intents (4 stages) & \(\text{SC} = \frac{\text{Number of covered intents}}{4}\) \\
\hline
\ac{L3 GR} & Proportion of teacher utterances using L3-level guidance & \(\text{L3 GR} = \frac{\text{Number of L3 guidance utterances}}{\text{Total teacher utterances}}\) \\
\hline
\ac{3C} & Success rate of correcting student errors to higher-order thinking & \(\text{\ac{3C}} = \frac{\text{Successful correction count}}{\text{Total error count}}\) \\
\hline
TotalScore & Weighted sum of all indicators & \(\text{TotalScore} = 0.15 \times \text{\ac{SD}} + 0.10 \times \text{\ac{SV}} + 0.15 \times \text{IKT} + 0.15 \times \text{BP} + 0.15 \times \text{SC} + 0.10 \times \text{L3 GR} + 0.20 \times \text{\ac{3C}}\) \\
\hline
\end{tabular}
}
\end{table*}

\subsubsection{Detailed Subjective Metrics}
To enable reliable and automated evaluation of interdisciplinary teacher-student dialogues, we adopt an LLM-as-a-Judge framework with a carefully crafted prompt to elicit consistent, high-quality structured scoring. The prompt is designed from the perspective of an educational expert, explicitly instructing the model to assess dialogues based on five defined dimensions of interdisciplinary teaching quality.

\subsection{Experiment Setup and Hyperparameters}
Table~\ref{tab:experiment} presents the hyperparameter settings for the large language model across different experimental setups. The temperature varies depending on the task, with a moderate value of 0.7 for multi-turn dialogue generation to balance creativity and coherence, while a lower temperature of 0.1 is used for subjective evaluation to ensure more deterministic outputs. The top-p parameter is applied selectively to control the diversity of generated text, and the maximum token length is adjusted to fit the expected output size for each experiment.
\begin{table}[!htbp]
  \small
  \centering
  \begin{tabular}{@{}lccc@{}}
    \toprule
    Experiments                      & Temperature & Top-P & Max Tokens \\
    \midrule
    Multi-turn Dialogue Gen.  & 0.7         & —     & 1000       \\
    Dialogue Annotation       & 0.8         & 0.95  & 5000       \\
    Subjective Evaluation     & 0.1         & 0.9   & 1024       \\
    \bottomrule
  \end{tabular}
  \caption{Hyperparameter Settings}
  \label{tab:experiment}
\end{table}

\subsection{Dialogue Generation Details}
We are building an interdisciplinary heuristic dialogue dataset for K-12 students, aiming to validate the guided teaching capabilities of Large Language Models (LLMs) in intelligent education scenarios. The dataset design is based on two sets of prompts: Student Prompts (Prompt 2) and Teacher Prompts (Prompt 3). The student-side prompt requires participants to choose any one of "five typical learning states" (does not understand the question's meaning, does not understand the explanation, has poor knowledge mastery, has weak curiosity, or has outstanding abilities). They then answer the teacher's questions in the first person, reflecting real thought processes or emotional responses, and demonstrating genuine learning difficulties and cognitive curves. The teacher-side prompt emphasizes the Socratic questioning method: all utterances are questions, complex propositions are broken down, and multiple disciplinary perspectives, including physics, biology, geography, and history, are considered. The teacher does not directly provide answers until the student achieves a deep-level understanding, at which point the teacher summarizes and marks "[End]" to close the dialogue.

Student prompts guide the model to simulate the thought processes of real learners by setting specific situations and roles; teacher prompts, on the other hand, focus on layered questioning strategies, guiding the model to break down difficult problems, explore their essence, and establish connections across multiple disciplinary dimensions. The two types of prompts work in conjunction to generate multi-turn dialogue instances, covering the complete cognitive path from initial confusion to knowledge internalization.

\begin{CJK*}{UTF8}{gbsn}
\begin{ch}{学生提示词}
n你是一名中学生，现在将模拟一次跨学科教学对话。请根据以下几种学生类型之一进行身份选择，并回应教师的提问。
可选情况：\\
（1）学生不理解问题的含义；\\
（2）学生不理解教师讲解的内容；\\
（3）某学生知识掌握较差；\\
（4）学生求知欲弱；\\
（5）学生各方面能力都很强。\\

请从中任选一个情况，作为学生进行回答。你的回答应遵循以下原则：
- 不要扮演老师；\\
- 回答必须与教师提问内容一致；\\
- 用第一人称表达；\\
- 展现出真实学生在该情境下的思维或情绪反应；\\
- 允许表达困惑、错误、兴趣或联系其他学科进行推理。\\

现在，请生成你的回应：
\end{ch}
\end{CJK*}

\subsection{Analysis}
\subsubsection{Overall Analysis}
Our experimental results reveal a profound and critical phenomenon: in the complex task of pedagogical guidance, there is a significant disconnect between a model's subjective fluency and its objective pedagogical effectiveness.
\begin{figure}[H]
  \centering
  \includegraphics[width=\columnwidth]{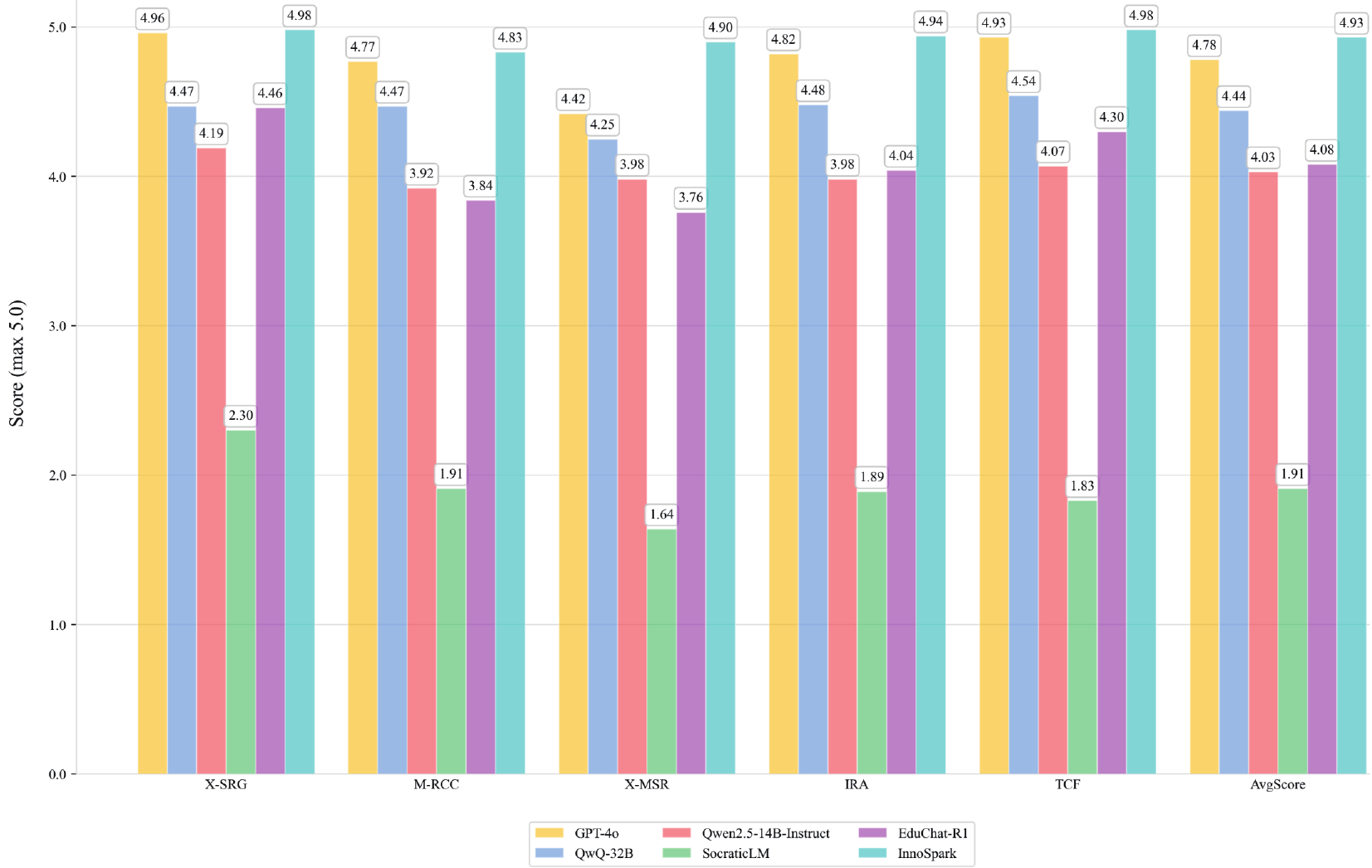}
  \caption{Visualization of model performance in subjective evaluation across key educational indicators.}
  \label{fig:subject_correlation}
\end{figure}
From the perspective of subjective Rubric scores (see Table \ref{fig:subject_correlation}), top-tier models like InnoSpark and GPT-4o perform nearly perfectly, with average scores (AvgScore) as high as 4.93 and 4.78, respectively. This is attributable to their excellent performance on metrics like Structural Integrity (SC) and L3 Guidance Rate (L3 GR), which enables them to generate structurally complete dialogues filled with open-ended questions that appear highly ideal. If we were to rely solely on this type of evaluation, we might conclude that current LLMs have largely solved the challenge of Socratic guidance.

\begin{figure}[H]
  \centering
  \includegraphics[width=\columnwidth]{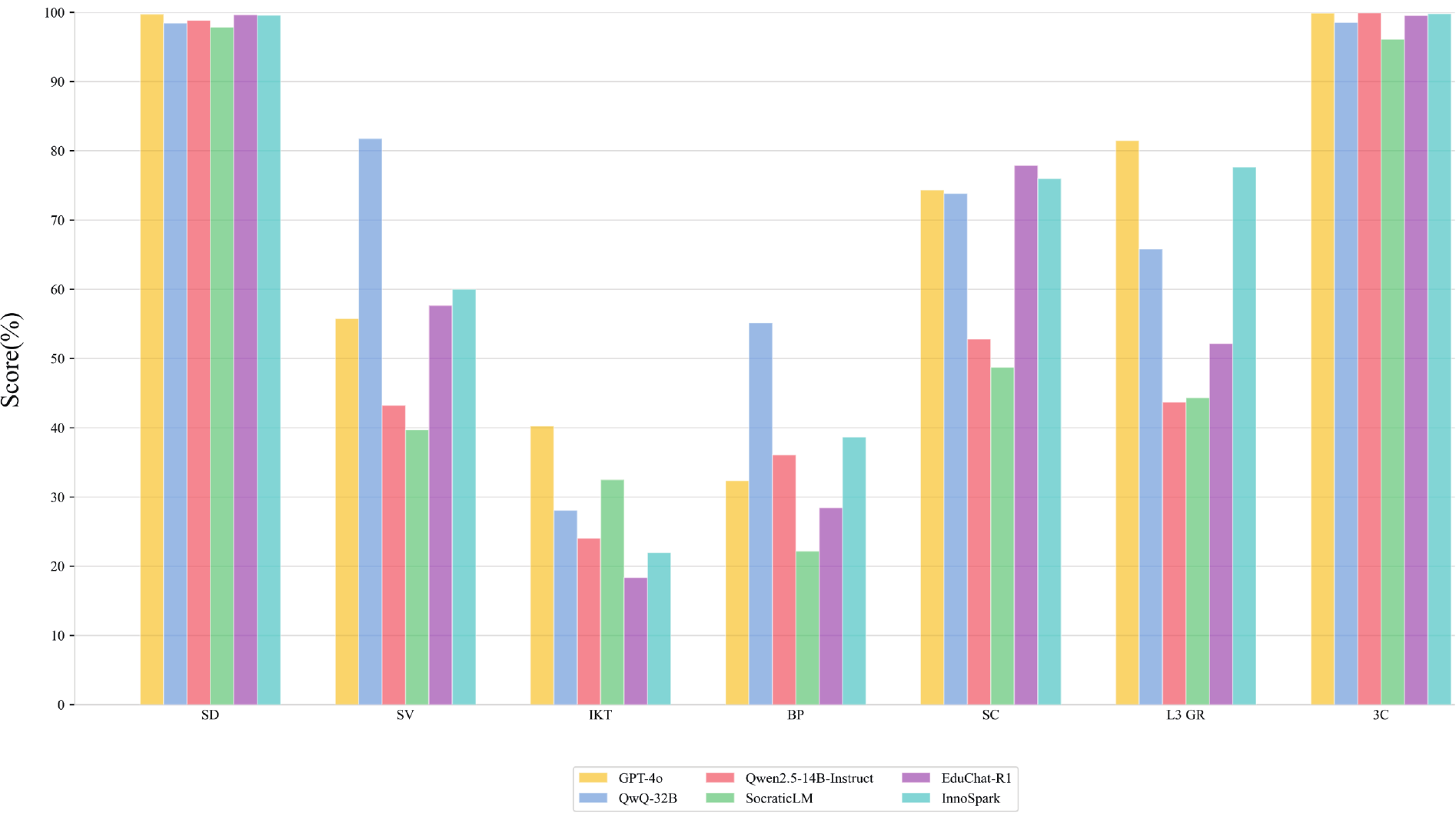}
  \caption{Visualization of model performance in objective evaluation across key educational indicators.}
  \label{fig:objective_correlation}
\end{figure}

However, our multi-dimensional objective metrics (see Table \ref{fig:objective_correlation}) unveil a deeper and more concerning truth. Current LLMs exhibit a widespread and severe deficiency on the two metrics that best represent the core value of our benchmark:
\begin{itemize}
    \item A collective failure in "Interdisciplinary Knowledge Transfer (IKT)": This is the weakest link across all models. Even GPT-4o, which scored highest on subjective ratings, only achieved an IKT score of 40.24\%, with most other models performing even worse. This provides strong evidence that while current LLMs are knowledgeable, they severely lack the ability to proactively guide students to make interdisciplinary connections. Their guidance is "vertically deep" rather than "horizontally integrative."

    \item A general mediocrity in "Bloom Progression (BP)": This is another key metric that exposes the models' shortcomings. Despite conducting multi-turn dialogues, the models were not effective in promoting a sustained increase in the students' cognitive levels. It is noteworthy that the best-performing model on the objective TotalScore, QwQ, achieved a relatively high BP score (55.17\%), yet its subjective ratings were not top-tier. This further corroborates the discrepancy between a model's "external performance" and its "internal pedagogical effect."
\end{itemize}

This series of findings clearly validates the argument we proposed in the introduction—"metrics serve as the conductor's baton for model optimization." If we only use traditional metrics focused on fluency and structure, we will be misled by the seemingly perfect performance of LLMs, thereby optimizing them into eloquent but ineffective "pseudo-tutors."

The core value of our SID benchmark lies precisely here: it is not merely a tool for "ranking" but rather a "diagnostic instrument." Through unique, process-and-outcome-oriented objective metrics like IKT and BP, it provides, for the first time, a "ruler" capable of revealing the true shortcomings of models in their deep pedagogical capabilities. This not only proves that current LLMs have a long way to go in achieving genuine interdisciplinary guidance but also points to the most critical optimization directions for developing future LLM systems that are truly pedagogically-aware and capable of fostering knowledge integration and transfer.

\subsubsection{Case Study}
\begin{figure*}[h!] 
 \centering \includegraphics[width=2.0\columnwidth]{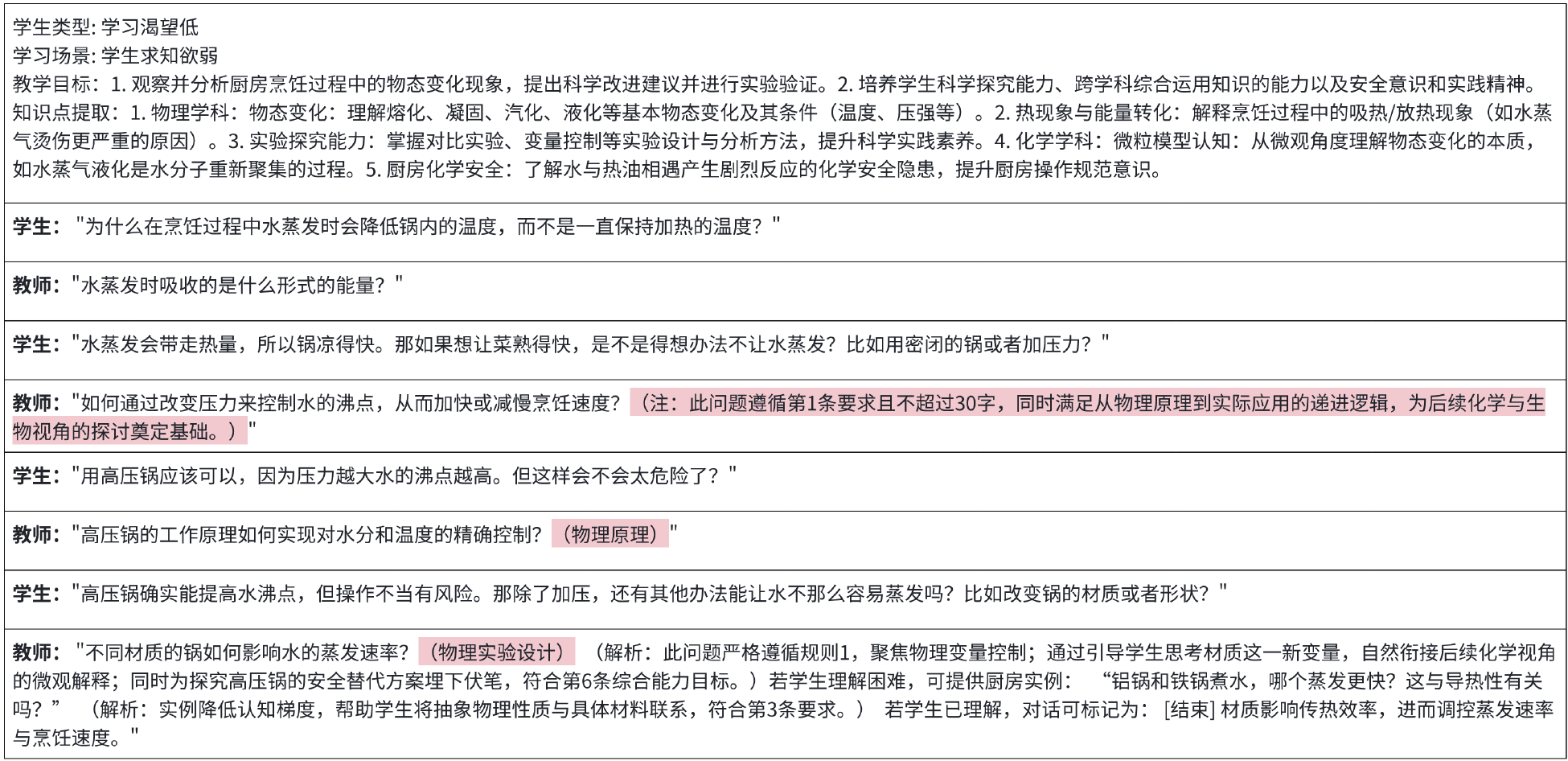} \caption{Case 1: Physics in Cooking (A Theory-Inquiry Task)(Chinese Version-InnoSpark)} \label{fig:case1_ch_inno} 
\end{figure*}

\begin{figure*}[h!] 
 \centering \includegraphics[width=2.0\columnwidth]{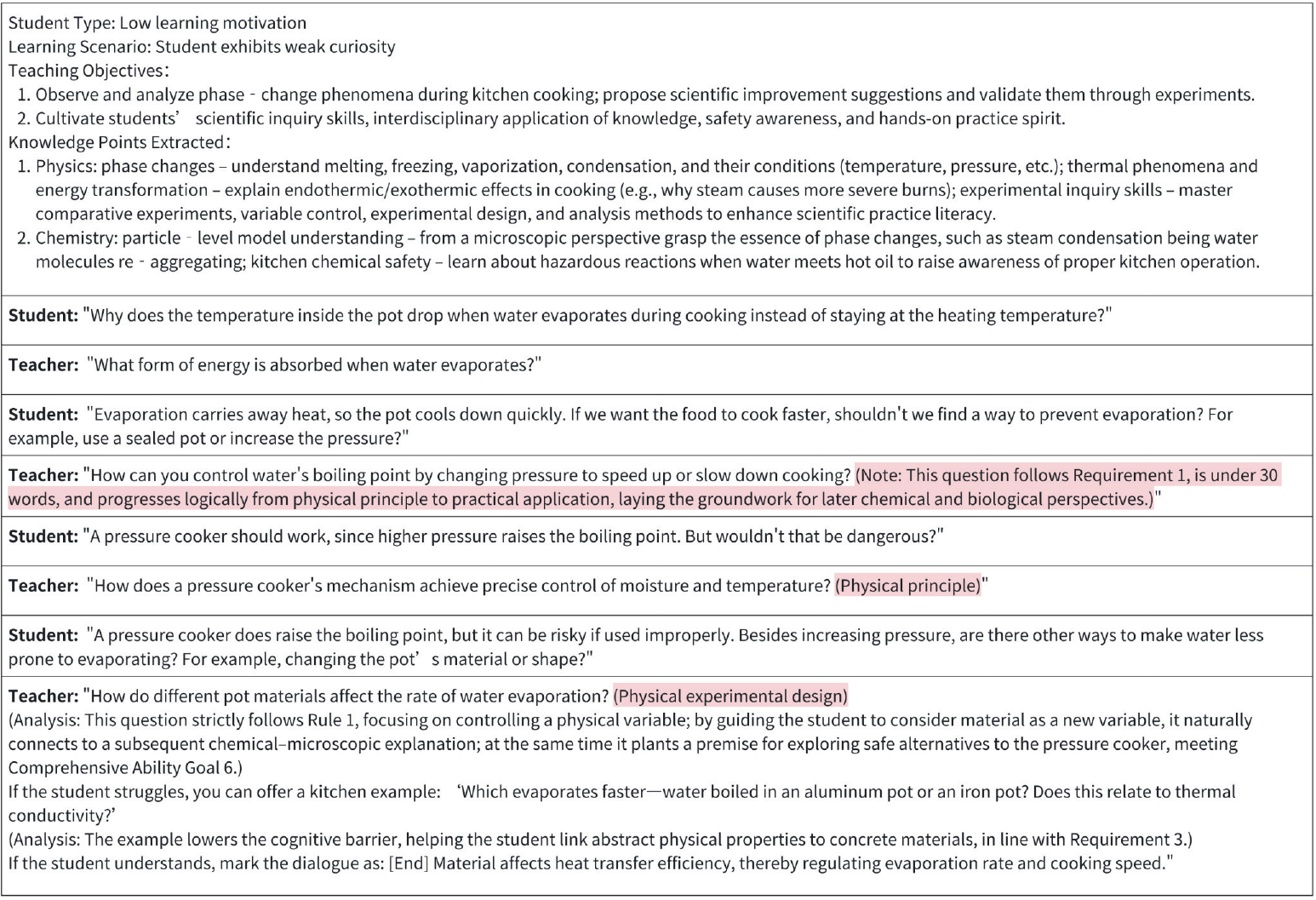} \caption{Case 1: Physics in Cooking (A Theory-Inquiry Task)(English Version-InnoSpark)} \label{fig:case1_en_inno} 
\end{figure*}

\begin{figure*}[h!] 
 \centering \includegraphics[width=2.0\columnwidth]{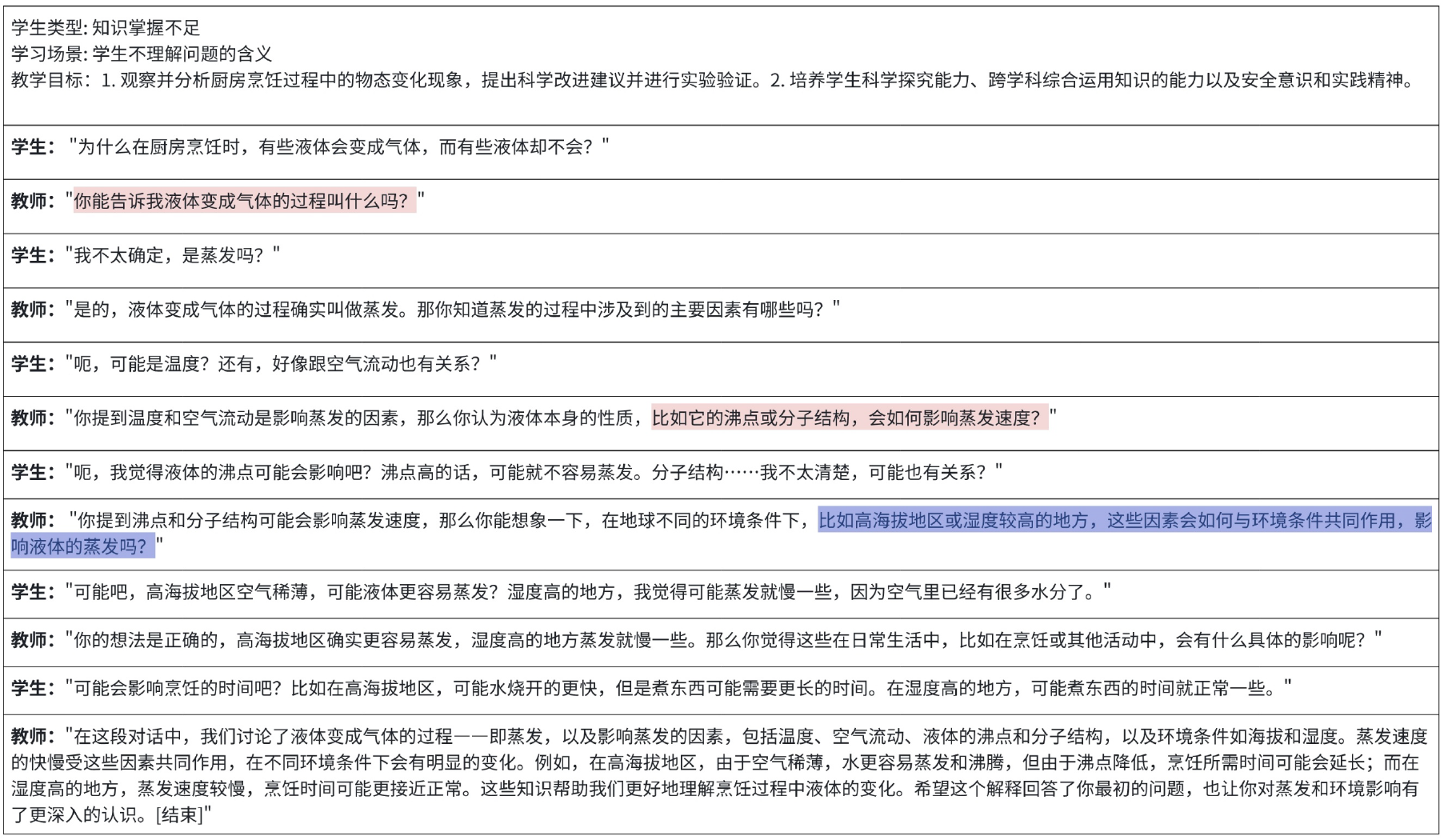} \caption{Case 1: Physics in Cooking (A Theory-Inquiry Task)(Chinese Version-GPT)} \label{fig:case1_ch_gpt} 
\end{figure*}

\begin{figure*}[h!] 
 \centering \includegraphics[width=2.0\columnwidth]{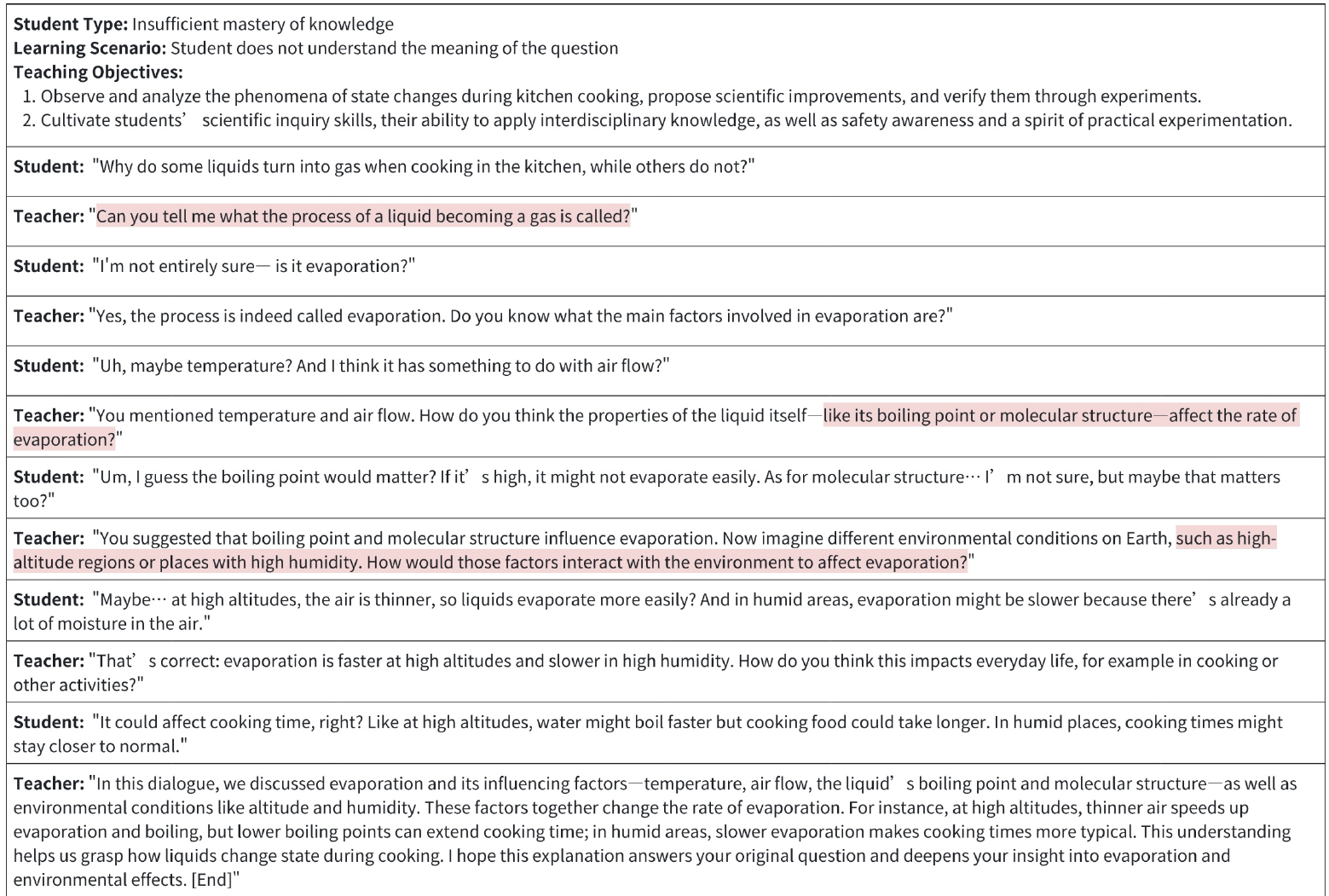} \caption{Case 1: Physics in Cooking (A Theory-Inquiry Task)(English Version-GPT)} \label{fig:case1_en_gpt} 
\end{figure*}

\begin{figure*}[h!] 
 \centering \includegraphics[width=2.0\columnwidth]{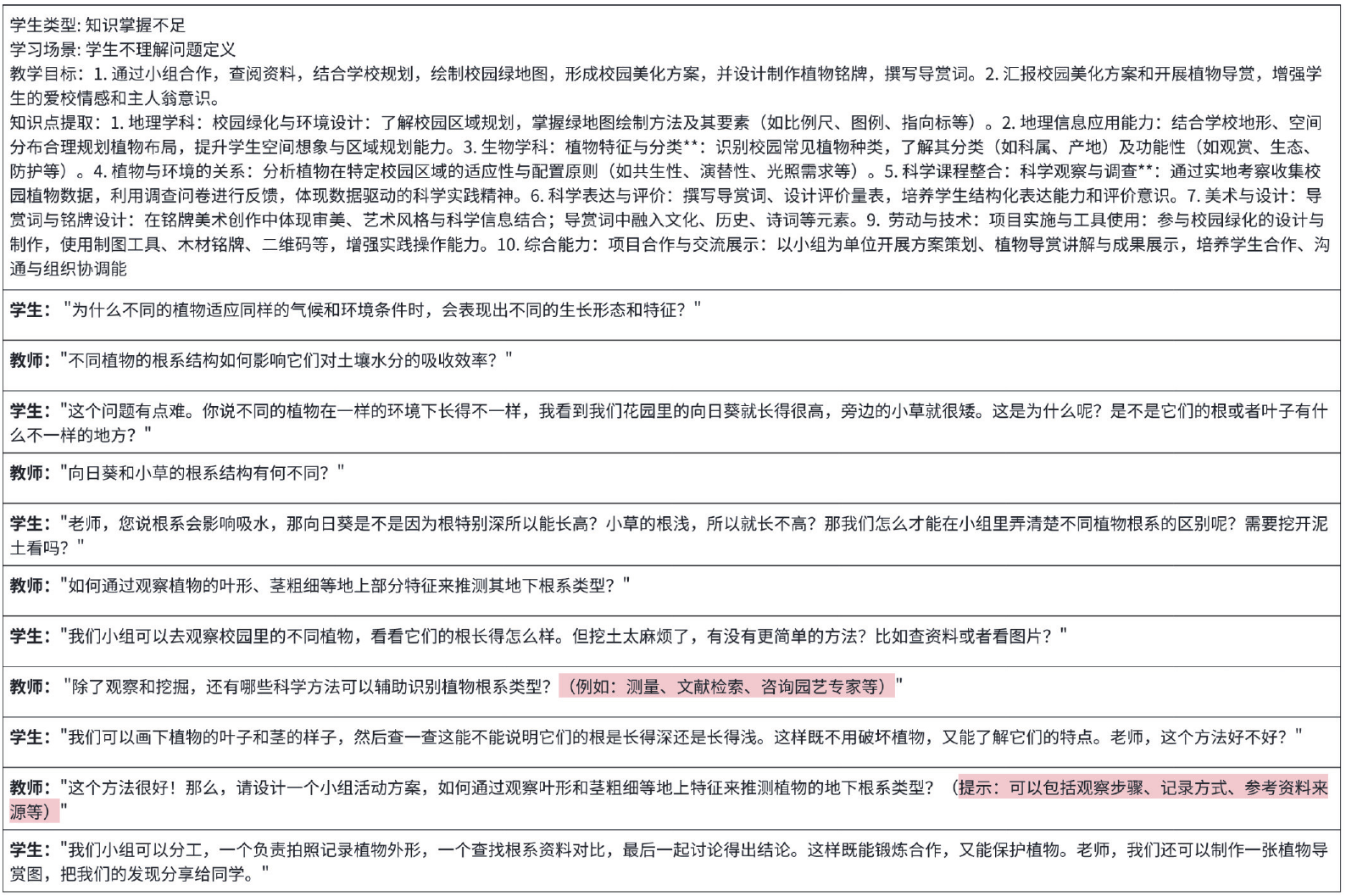} \caption{Case 2: Campus Ecological Beautification (A Practical Design Task)(Chinese Version-InnoSpark)} \label{fig:case2_ch_inno} 
\end{figure*}

\begin{figure*}[h!] 
 \centering \includegraphics[width=2.0\columnwidth]{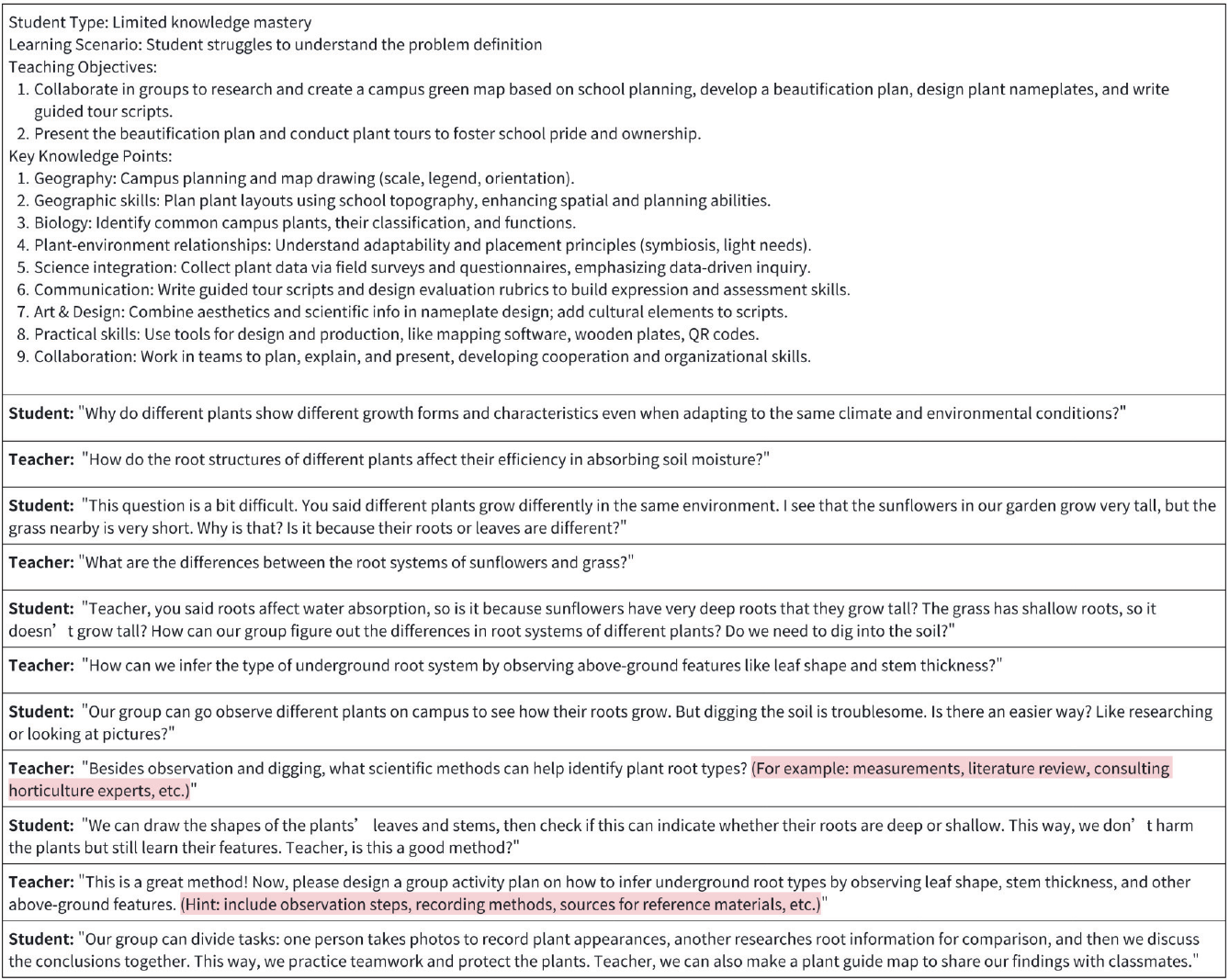} \caption{Case 2: Campus Ecological Beautification (A Practical Design Task)(English Version-InnoSpark)} \label{fig:case2_en_inno} 
\end{figure*}

\begin{figure*}[h!] 
 \centering \includegraphics[width=2.0\columnwidth]{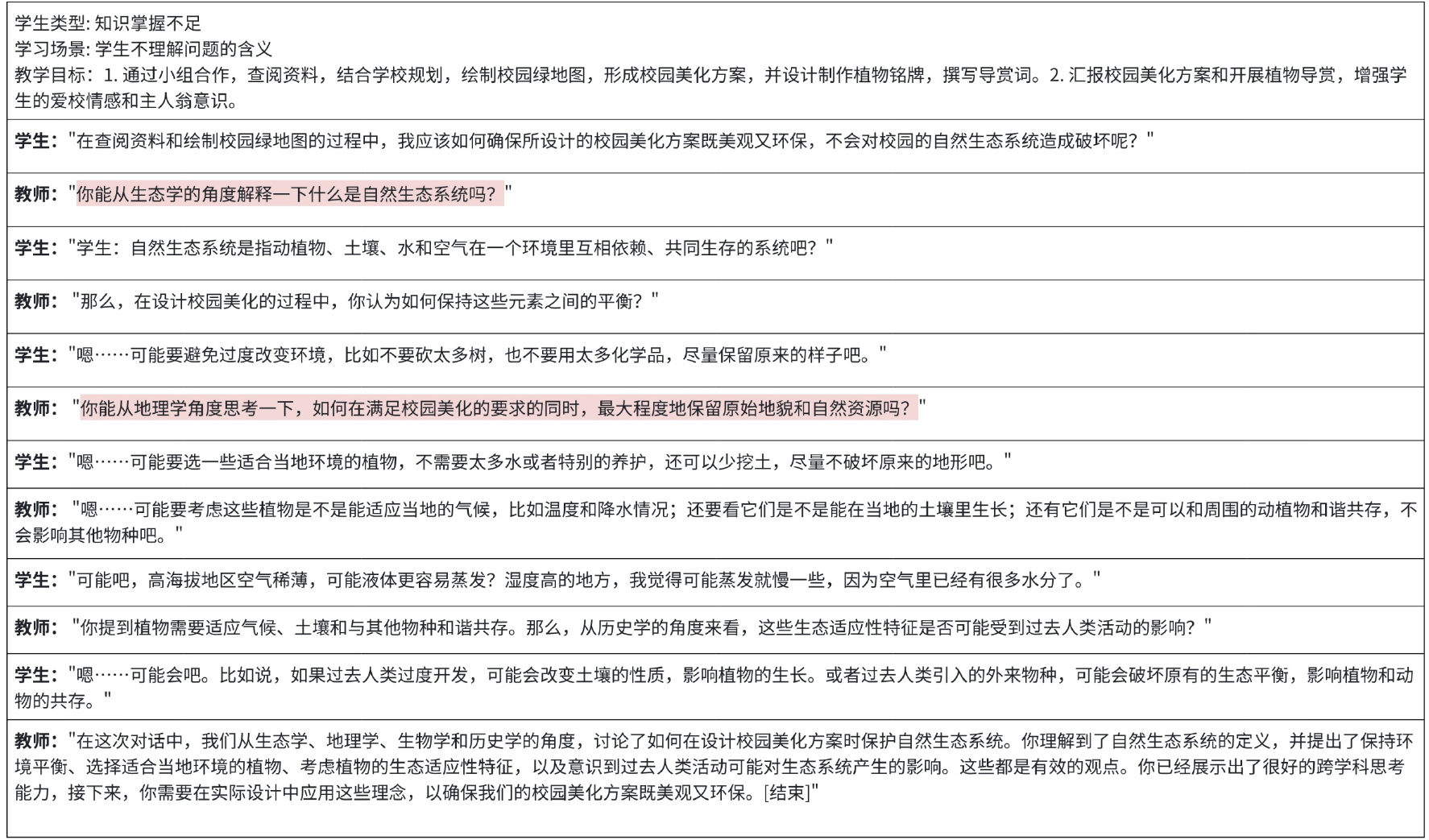} \caption{Case 2: Campus Ecological Beautification (A Practical Design Task)(Chinese Version-GPT)} \label{fig:case2_ch_gpt} 
\end{figure*}

\begin{figure*}[h!] 
 \centering \includegraphics[width=2.0\columnwidth]{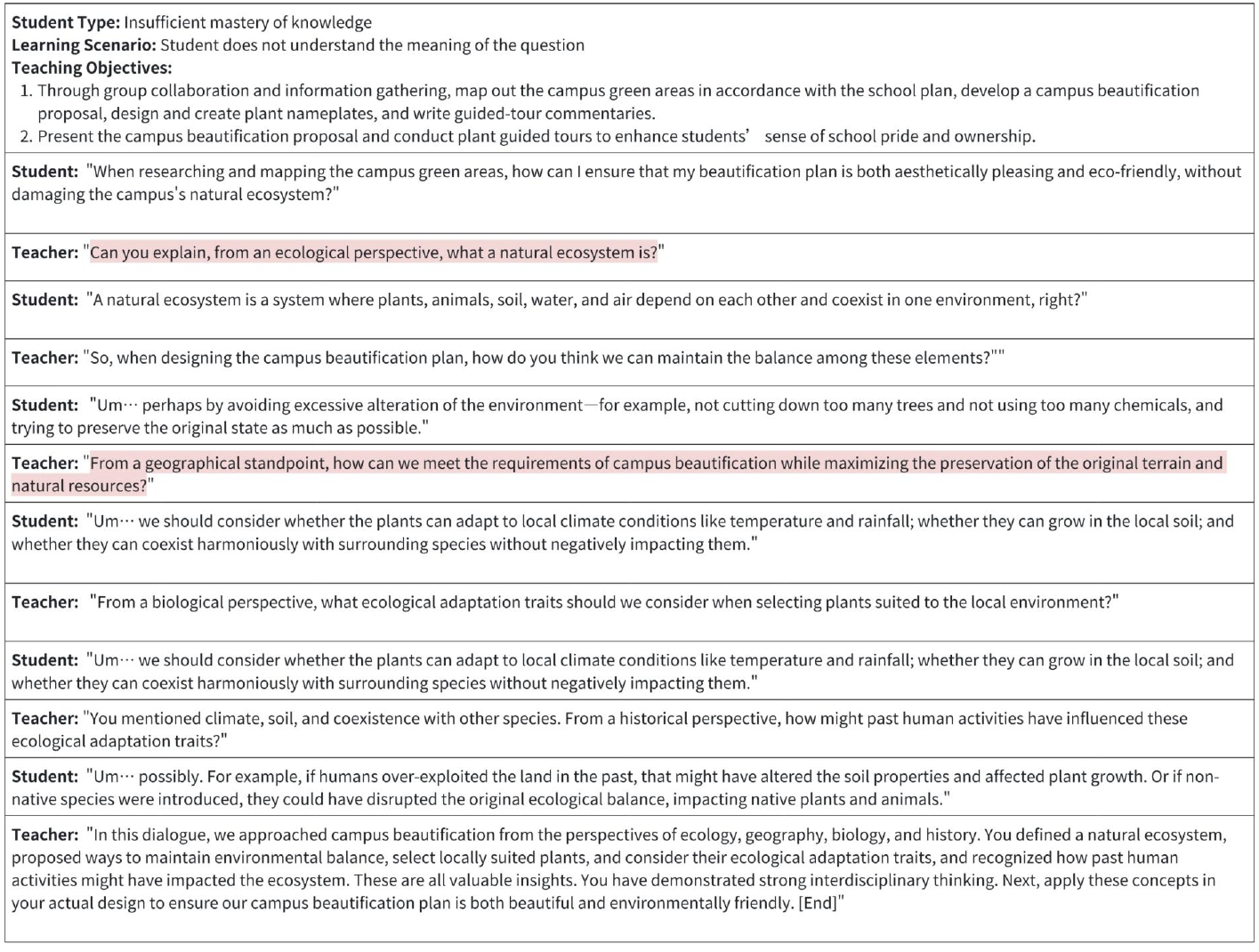} \caption{Case 2: Campus Ecological Beautification (A Practical Design Task)(English Version-GPT)} \label{fig:case2_en_gpt} 
\end{figure*}

\begin{figure*}[h!] 
 \centering \includegraphics[width=2.0\columnwidth]{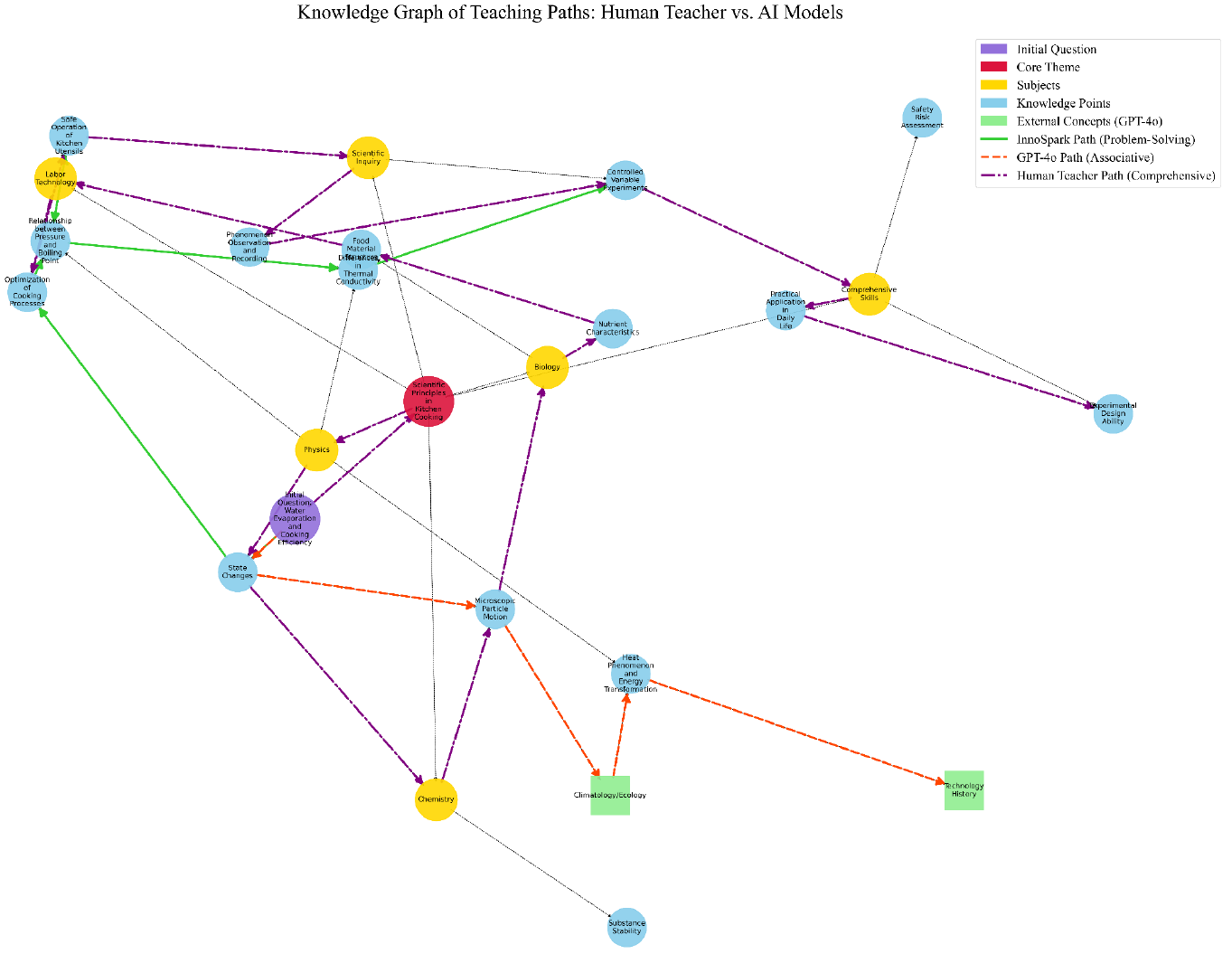} \caption{Interdisciplinary Case Study Knowledge Map: Conceptual Framework and Connections} \label{fig:knowledge} 
\end{figure*}

To provide a nuanced, qualitative analysis of current model capabilities, we conducted a comparative case study covering multiple pedagogical topics, with detailed dialogues shown in Figure \ref{fig:case1_ch_inno}-\ref{fig:case2_en_gpt}. We analyzed the guided dialogues of a general-purpose LLM (GPT-4o) and a specialized educational LLM (InnoSpark) on two different interdisciplinary STEM pedagogical tasks—"Physics in Cooking" and "Campus Ecological Beautification"—and contrasted their performance with the guidance strategies of an ideal human expert. Concurrently, we visualized the disciplines and interdisciplinary knowledge points involved in both pedagogical tasks to more clearly illustrate the teacher's interdisciplinary guidance capabilities, as shown in Figure \ref{fig:knowledge}.

\noindent \textbf{Model Capability Profiles: Different Strategic Preferences and Common Bottlenecks}

Our analysis reveals that different LLMs exhibit distinct strategic preferences when faced with different types of pedagogical tasks, yet they also share profound bottlenecks.

\noindent Case 1: Physics in Cooking (A Theory-Inquiry Task)

In this more theory-oriented inquiry task, the two models performed as follows:
\begin{itemize}
    \item General-Purpose Large Language Model (GPT-4o): As shown by the orange dotted path in the figure, GPT-4o's guidance is a divergent, breadth-first exploration. Starting from the initial question, after covering core physics concepts like state changes and microscopic particle motion, it quickly performs two large-scale analogical transfers: one towards the external macro-concepts of climatology/ecology, and another towards technology history. The strength of this strategy lies in its exceptional knowledge breadth, which can demonstrate the universality of scientific principles to the student. However, this comes at the cost of sacrificing the authenticity of the pedagogical context, and its guidance path more closely resembles a teacher-led, divergent "conceptual tour."

    \item Specialized Educational Large Language Model (InnoSpark): In stark contrast, as shown by the green solid path in the figure, InnoSpark's guidance is a convergent, depth-first inquiry. Its dialogue remains tightly coupled with the cooking scenario, focusing on solving concrete problems. Its path clearly covers concepts from state changes to the relationship between pressure and boiling point, then to the safe operation of kitchen utensils, and finally lands on differences in thermal conductivity and controlled variable experiments. This path perfectly embodies the pedagogical goal of "Scientific Inquiry and Engineering Practice", but its weakness is its incomplete structure (the dialogue ends abruptly), and its complete confinement within physics, failing to achieve any interdisciplinary knowledge transfer.
\end{itemize}

\noindent Case 2: Campus Ecological Beautification (A Practical Design Task)

In this more practice-oriented design task, the models' strategic preferences shifted:
\begin{itemize}
    \item General-Purpose Large Language Model (GPT-4o): Faced with a design task, GPT-4o adopted a structured, checklist-style of interdisciplinary questioning, guiding the student to think from the perspectives of ecology, geography, biology, and history in sequence. Its strength is its broad coverage, but its weakness is that the interdisciplinary connections feel rigid and teacher-led.

    \item Specialized Educational Large Language Model (InnoSpark): InnoSpark employed a "process of elimination" as a structured scaffold, breaking the complex problem into a series of manageable steps. More importantly, it successfully identified and intervened in a student's misconception, but its method of correction was non-Socratic ("direct negation" and "knowledge infusion").
\end{itemize}

Comparative Analysis and The Gap with Human Teachers
This cross-case comparison reveals the common deficiencies of current LLM tutors:
\begin{itemize}
    \item Lack of Dynamic Strategy Switching: No LLM was able to flexibly switch between different guidance styles (e.g., from "associator" to "decomposer") based on the task type (theory inquiry vs. practical design) or the student's real-time responses.

    \item Insufficient Error Handling Capabilities: Whether completely ignoring errors (GPT-4o) or performing a clumsy correction (InnoSpark), the LLMs failed to demonstrate the expert teacher's advanced ability to use misconceptions as teachable moments.

    \item Passive and Rigid Interdisciplinary Guidance: The LLMs' interdisciplinary connections were either a teacher-led "checklist" tour, completely absent, or reliant on the student's spontaneous transfer. They universally lacked the creativity for proactive and organic interdisciplinary guidance.
\end{itemize}

This cross-case comparison reveals a common deficiency among current LLM tutors: whether it is GPT-4o's divergent association or InnoSpark's convergent decomposition, their guidance paths exhibit a preset and linear nature. They universally lack the capacity for dynamic strategy switching, diagnosing and leveraging student errors, and proactive, organic interdisciplinary guidance.

When compared to an expert human teacher, all AI models pale in comparison. As illustrated by the purple path in the knowledge graph, a human teacher's guidance is a dynamic, network-like process of cognitive construction. It not only covers core physics knowledge such as `state changes` and the `relationship between pressure and boiling point`, but also, when discussing practical applications like pressure cookers, it timely and adaptively steers the dialogue towards the `Comprehensive Skills` dimension of `Safety Risk Assessment` and the `Scientific Inquiry` dimension of `Experimental Design Ability`. This guidance does not follow a fixed script but instead establishes meaningful, non-linear connections between different knowledge nodes based on the student's responses. Most critically, a human teacher can flexibly switch between different guidance strategies based on real-time student feedback and, finally, release the scaffolding to guide the student toward self-constructed summarization.

This cross-case study demonstrates that while LLM has become considerably capable of replicating the "form" of structured guidance—able to execute different strategies like "logical deduction", "disciplinary tours", or "problem decomposition"—a fundamental gap remains in capturing the "essence" of teaching. This essence lies in the dynamic adaptability to handle pedagogical contingencies, the interdisciplinary creativity to foster knowledge integration, and the deep diagnosis of students' cognitive processes. This strongly validates the need for developing a benchmark like SID, which is capable of capturing and evaluating these deep, multi-dimensional pedagogical capabilities, to drive the field of LLMs in education forward.
\begin{prm}{Prompt of Teacher}
yYou are an interdisciplinary instructor who always employs the Socratic method to guide your students. Your aim is not to provide answers directly, but to help students think for themselves and build interdisciplinary understanding through a series of progressively deeper questions.
Your code of conduct is as follows:

Every one of your utterances must take the form of a question.\\
You must never provide answers outright; you may only guide students’ thinking through questions.\\
You should break complex problems down into smaller, simpler questions until the student can understand and answer them.\\
All questions should integrate knowledge across multiple disciplines, prompting students to make connections in physics, biology, geography, history, and more.\\
You should constantly probe with "Why?”, "How do you know?”, "What is the connection to other disciplines?”, etc., to drive deeper conceptual understanding.\\
If you determine that the student has achieved full understanding, please conclude your response with a brief summary and mark it: [End]
\end{prm}

\begin{prm}{Prompt of Student}
yYou are now a middle-school student taking part in a simulated interdisciplinary teaching dialogue. Please choose one of the following student types and respond to the teacher’s question accordingly:

Options:

The student does not understand what the question means.\\
The student does not understand the teacher’s explanation.\\
The student has a weak grasp of the subject matter.\\
The student lacks curiosity.\\
The student excels in all areas.\\

Select one of these scenarios and reply as the student. Your response should adhere to the following guidelines:

Do not play the role of the teacher;\\
Ensure your answer directly addresses the teacher’s question;\\
Speak in the first person;\\
Convey the genuine thoughts or emotions a student in that situation would have;\\
Feel free to express confusion, mistakes, interest, or make connections to other subjects in your reasoning.\\

Now, please generate your response.
\end{prm}


\begin{figure*}[h!] 
 \centering \includegraphics[width=2.0\columnwidth]{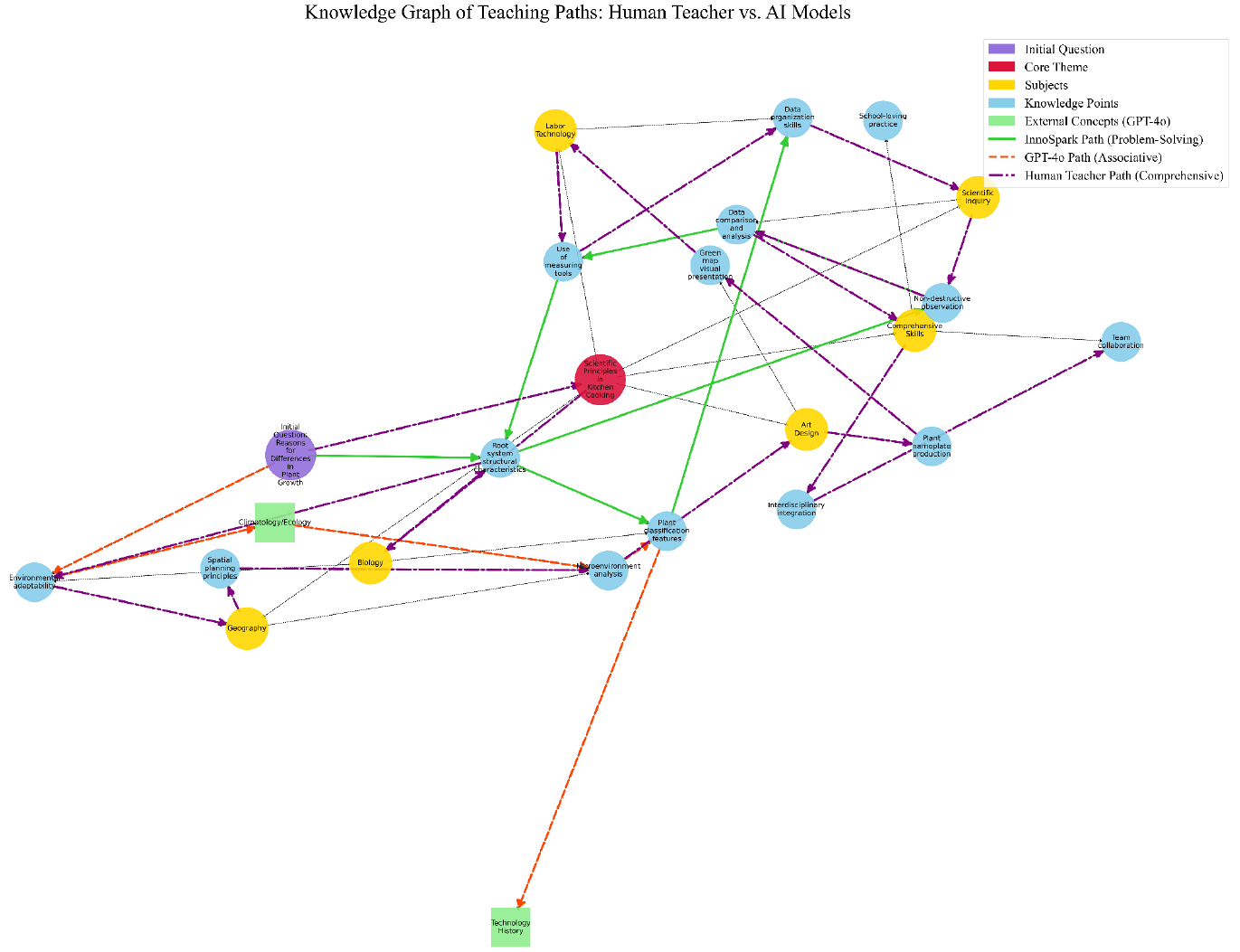} \caption{Interdisciplinary Case Study Knowledge Map: Conceptual Framework and Connections} \label{fig:knowledge2} 
\end{figure*}

\subsection{Appendix F: Future Work}
Our current work successfully establishes a benchmark capable of evaluating the capabilities of LLMs in complex, interdisciplinary guidance. However, the student models used in our dialogue generation are primarily based on predefined student types and pedagogical scenarios, which simplifies the complexity of real learners to some extent. Therefore, our core future research direction is to develop a more dynamic, multi-dimensional, and deeply contextualized Learner Model to drive our dialogue generation framework.
Specifically, we plan to deepen our research in the following two areas:

\noindent \textbf{1. Expanding the Dimensions of Learner Modeling}

To make the simulated students more realistic, we will introduce richer, dynamically changing dimensions of learner modeling on top of our existing "student cognitive state" annotation. This will not only generate more challenging dialogues but also provide a more fine-grained metric for evaluating the personalized adaptive capabilities of the LLM teacher. The key fields we plan to introduce include:
\begin{itemize}
\item Cognitive Dynamics: Beyond just the static level of knowledge mastery, we aim to model the student's learning speed and style (student\_learning\_speed) to differentiate between students who require step-by-step progression and those capable of making conceptual leaps.

\item Affective State: We will introduce the modeling of affective states (student\_emotional\_state), such as confidence, anxiety, or hesitation. This is crucial for evaluating the LLM teacher's ability to provide emotional support and motivational encouragement.

\item Metacognition and Strategies: We will add modeling for metacognitive abilities (student\_metacognition) and learning strategy preferences (student\_strategy\_preference) to simulate more complex learning behaviors, such as whether a student facing difficulty will actively seek verification, prefer analogies, or rely on the teacher.
\end{itemize}

\noindent \textbf{2. Enriching the Pedagogical Context for Dialogue Generation}

To make the pedagogical dialogues more authentic and ecologically valid, we will configure a richer, more structured Pedagogical Context for each dialogue generation instance. This will enable our benchmark to not only evaluate the general guidance capabilities of a LLM but also its performance under specific instructional constraints. The modules we plan to introduce include:
\begin{itemize}

\item Curriculum Anchoring: Explicitly defining the specific curriculum standards and grade level that each dialogue is based on.

\item Defining Prior Knowledge: Explicitly defining the prior knowledge of the simulated student to generate more targeted dialogues that occur within the student's Zone of Proximal Development (ZPD).

\item Knowledge Graph Alignment: Aligning the dialogue's objective with a node or path in a structured knowledge graph to provide the LLM teacher with more precise pedagogical goals and knowledge boundaries.

\item Comprehensive Student Profile: Finally, integrating all the aforementioned dimensions and contextual information into a comprehensive Student Profile that will serve as the initial input for dialogue generation.
\end{itemize}

By introducing this more complex and dynamic learner model, we expect to generate dialogues that are more authentic, more challenging, and more diverse. This will not only greatly enrich our SID dataset but also push the entire research community to develop and evaluate the next generation of truly personalized LLM pedagogical agents that understand not only teaching strategies but also the students themselves.

\end{document}

%% file: 01.Introduction.tex
\IEEEraisesectionheading{\section{Introduction}\label{sec:introduction}}
\begin{figure*}[ht]
\begin{center}
\centerline{\includegraphics[width=1.0\textwidth]{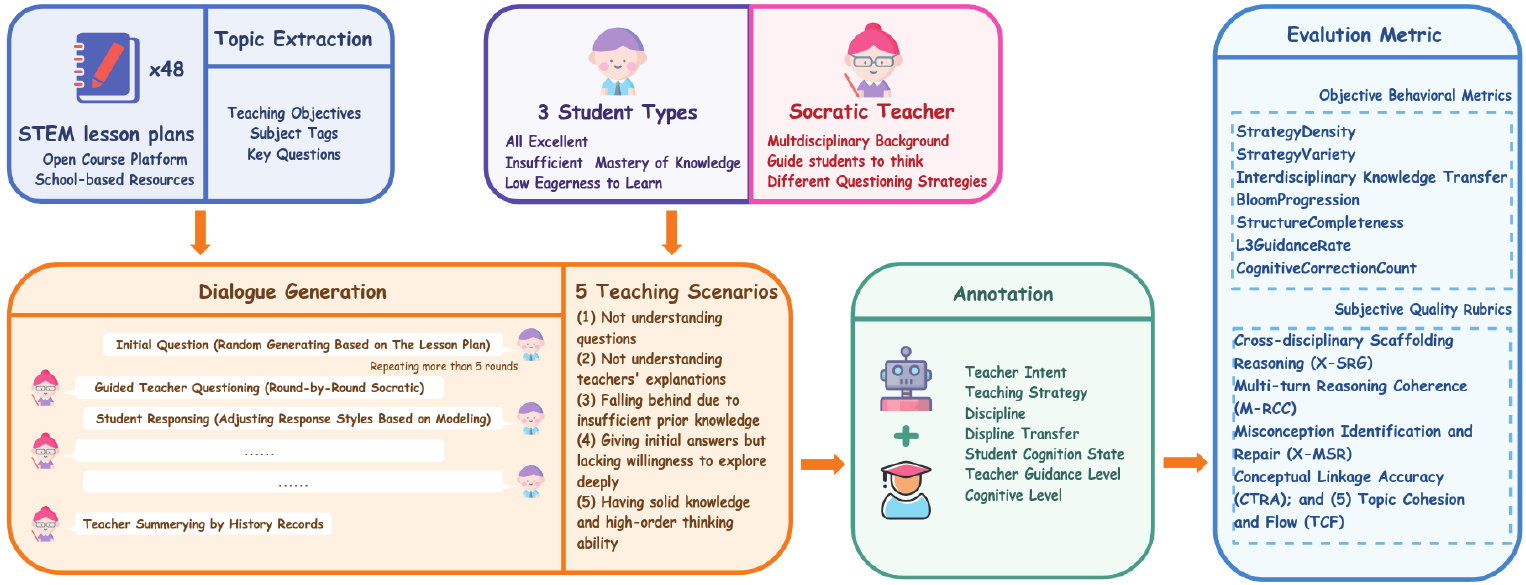}}
\vspace{-0.2cm}
\caption{An overview of the SID construction and evaluation framework. The process begins with topic extraction from 48 real-world STEM lesson plans, which seeds a multi-turn dialogue generation process between a "Socratic Teacher" agent and "Student" agent personas. The resulting dialogues are then annotated using a detailed, multi-dimensional pedagogical schema, which in turn provides the basis for our two-tiered (objective metrics \& subjective rubrics) evaluation framework.}
\label{fig:framework}
\end{center}
\vspace{-0.6cm}
\end{figure*}
\noindent In modern education, cultivating students' ability for knowledge integration and transfer in complex contexts is a core objective, for which interdisciplinary STEM education serves as a key pathway~\cite{vilsmaier2015case}. However, this form of teaching, which relies on high-level expert guidance, is exceptionally difficult to scale~\cite{santos2017interdisciplinarity}. \ac{LLMs}, with their vast knowledge and flexibility, offer immense potential to solve this scaling challenge~\cite{cai2024systemic}, a core question arises: how can we scientifically evaluate whether LLMs have truly acquired the capability to guide students toward knowledge integration and transfer? The current evaluation ecosystem exhibits a critical gap on this key task, leaving us unable to measure the true performance of LLMs in complex guidance scenarios.

To scientifically address this question, we must first define what core, evaluable capabilities a "good" guide must possess, a definition that must be grounded in robust pedagogical and cognitive science theories. Constructivism emphasizes that learning is a process of active meaning-making by the student, rather than passive reception of information~\cite{fosnot1996constructivism, hein1991constructivist}. This mandates that LLMs must transcend static knowledge-base Q\&A and instead support knowledge application in authentic tasks. More critically, the \ac{ZPD} theory provides the theoretical foundation for LLM's pedagogical intervention: effective instruction occurs at the boundary of a student's independent capabilities, facilitating cognitive advancement through adaptive pedagogical scaffolding. This requires LLMs to possess two core capabilities which include inferring the student's latent cognitive state and dynamically generating guiding strategies~\cite{chaiklin2003zone}.

To operationalize the aforementioned capabilities, the Socratic method provides an ideal framework. It is not only an empirically validated strategy for promoting deep conceptual learning but, more critically, its inherent structure provides an excellent entry point for computational modeling~\cite{hintikka2004socratic}. Unlike fully divergent, open-ended conversations, Socratic dialogue adheres to a series of identifiable pedagogical strategies. Through a coherent and progressive process of questioning, it guides students to clarify ideas, challenge assumptions, and make analogical transfers, thereby fostering not only the gradual construction of systematic cognitive structures but also the critical ability to transfer knowledge to new contexts. Simultaneously, a teacher can guide students through a pathway of "identifying conflicts, comparing concepts, reorganizing knowledge, and synthesizing understanding" to facilitate the acquisition of knowledge integration skills~\cite{bimantara2025cognitive}. Crucially, it is this structured nature that makes it possible to formalize and model the intrinsic pedagogical intents, strategic sequences, and cognitive impacts within the teaching process, rendering it an ideal and measurable object for evaluation~\cite{davoudi2015systematic}.

However, the current evaluation ecosystem in the field of LLM education has a critical gap in measuring whether LLMs can implement the aforementioned complex Socratic guidance. Firstly, mainstream teaching dialogue datasets are mostly limited to task-oriented corpora that are single-disciplinary, short-turn, and weakly structured \cite{liu2023m3ke, gu2024xiezhi}. These materials themselves lack samples that reflect in-depth exploration and strategic guidance, so they are insufficient for evaluating LLMs' ability in complex multi-turn interdisciplinary dialogues. This deficiency is even more prominent in evaluation metrics. Existing frameworks mostly rely on superficial indicators such as fluency or answer accuracy \cite{tanaka2024leveraging}, failing to conduct fine-grained, process-oriented evaluations of the teaching quality of dialogues. However, metrics are the baton of models; the lack of such in-depth evaluations may mislead LLMs into becoming fluent and seemingly helpful chatbots, which are unable to better support students' abilities in knowledge integration and transfer. 
The lack of rigorous evaluation risks promoting LLMs that appear fluent but fail to foster core skills like knowledge integration and transfer.
Without a benchmark that can measure these deeper skills, the development of genuinely effective, LLM-based pedagogical agents will remain stalled.

To address these gaps in evaluation, this paper introduces \ac{SID} (Fig. \ref{fig:framework}), a new benchmark for Socratic pedagogical dialogues centered on interdisciplinary STEM tasks. Our main contributions are threefold:

\begin{itemize}
\item \textbf{A novel benchmark dataset grounded in educational theories for interdisciplinary STEM dialogue.} Our dataset, SID, is a large-scale benchmark comprising more than 10,000 dialogue turns across 48 complex STEM projects. 
The entire design aims to reflect authentic pedagogical interactions and cognitive progressions, grounded in theories of Constructivism and \ac{ZPD}.
\item \textbf{A structured annotation schema and semi-automated pipeline for deep pedagogical analysis.} We design a nine-field schema covering pedagogical intents, strategies, and knowledge transfer, providing an operational foundation for building and evaluating pedagogically-aware intelligent agents.
\item \textbf{A suite of new evaluation tasks and metrics that reveal significant bottlenecks in current \ac{LLMs}.} We propose novel metrics such as \ac{X-SRG} and \ac{M-RCC}. Our experiments demonstrate that even state-of-the-art LLMs (e.g., GPT-4o) perform poorly on our benchmark, validating its challenge and providing an empirical basis for future model optimization.
\end{itemize}

%% file: 02.Related_works.tex
\section{Related Work}
\subsection{AI for Socratic Interdisciplinary STEM Education}
\noindent AI is increasingly applied to solve the core challenges of STEM education. Early explorations in \ac{AIEd}, such as \ac{ITS} for structured problem-solving~\cite{lin2023artificial,stamper2024enhancing}, achieved success in single disciplines but their rigid, rule-based architectures are incapable of handling the complex, ill-structured nature of Socratic, interdisciplinary inquiry tasks. The advent of \ac{LLMs}, with their vast knowledge and flexibility, appeared to open a new frontier for solving this problem~\cite{wang2025llm,piro2024mylearningtalk}. Researchers have explored their potential along two primary lines: (1) leveraging general-purpose LLMs (e.g., GPT-4) for broad interdisciplinary assistance~\cite{achiam2023gpt}, and (2) fine-tuning specialized models (e.g., SocraticLM, EduChat) on instructional data to enhance their pedagogical capabilities~\cite{liu2024socraticlm,dan2023educhat}.

However, regardless of the approach taken, these LLM-based solutions share a common limitation at their pedagogical core, as their dialogue strategies often aim to provide direct support or probe comprehension of a single text rather than scaffold the student's own reasoning process in complex, multi-turn, interdisciplinary contexts~\cite{wu2025collabllm}. This widespread deficiency stems from the lack of effective guidance throughout the model development ecosystem, since there are no benchmarks available to evaluate their true pedagogical capabilities. Therefore, creating a benchmark that faithfully represents and evaluates the complexity of multi-turn, Socratic, and interdisciplinary dialogue is the critical next step to advance the field.
\subsection{Benchmark for Interdisciplinary Socratic Dialogue}
\noindent The construction of high-quality benchmarks is foundational to advancing the field, yet a critical review of the existing landscape reveals a persistent, multifaceted gap in both Socratic and interdisciplinary dimensions. Early methods for creating pedagogical dialogues, whether relying on predefined rules (e.g., in AutoTutor~\cite{graesser2001intelligent}) or crowdsourcing (e.g., CIMA~\cite{stasaski2020cima}), were limited by poor scalability and an inability to capture the complexity of authentic inquiry. Even with the advent of LLM-assisted generation, recent datasets like MATHDIAL and Book2Dial~\cite{macina2023mathdial, wang2024book2dial} still focus on single subjects, failing to address the challenge of interdisciplinary knowledge synthesis. Meanwhile, benchmarks that focus more on pedagogical strategy also exhibit critical shortcomings. For instance, SocraticLM~\cite{liu2024socraticlm} models Socratic tactics but is aimed at probing comprehension of a single text, not facilitating knowledge synthesis between different scientific fields. M3KE~\cite{liu2023m3ke}, while targeting multi-step reasoning, remains within a single domain, failing to capture the holistic planning required for complex STEM projects. In summary, existing benchmarks either focus on overly simplified, single-domain tasks or fail to effectively integrate pedagogical strategy with interdisciplinary reasoning. This leaves a critical gap unaddressed, which our work, SID, is designed to fill.

%% file: 03.Benchmark.tex
\section{SID Benchmark \label{sec:benchmark}}
\subsection{Design Principles}
\noindent To advance the application of \ac{LLMs} in education, particularly for interdisciplinary pedagogical dialogue, we construct the \ac{SID} benchmark. To ensure a rich diversity of interactions, the dialogues for each of the 48 projects are generated by simulating 20 distinct students. These student profiles are systematically created by sampling from 3 core student types and placing them within 5 typical pedagogical scenarios. With each simulated student initiating two distinct lines of inquiry, this results in a total of 1,920 unique, goal-oriented dialogues. Each dialogue is required to have a minimum of five turns of teacher-student interaction to ensure sufficient depth for cognitive progression. This benchmark is designed to provide a standardized evaluation framework for assessing LLM capabilities in guided cross-disciplinary conversations, with a special focus on key challenges such as cross-domain knowledge transfer, strategic pedagogical guidance, student cognitive state modeling, and instructional intent tracking. The entire design is grounded in Constructivist learning theory and \ac{ZPD} theory, operationalized through the principles of Socratic questioning to ensure the dialogues possess sufficient interdisciplinary coverage, cognitive depth, and pedagogical diversity.
\subsection{Dataset Construction}
\noindent \textbf{Data collection} The foundation of this benchmark is a set of meticulously curated interdisciplinary STEM lesson plans grounded in the Chinese educational context, resulting in a high-quality, Chinese-language dialogue dataset. Our construction process begins with a broad screening of public Chinese educational resources, primarily including the Chinese National Curriculum Standards, mainstream K-12 textbooks, and online teaching platforms. From this pool, we distill initial pedagogical tasks that necessitate the synthesis of knowledge from different fields to solve a practical problem. Subsequently, these selected tasks undergo a process of secondary development and standardization by our team, ultimately resulting in 48 unique lesson plans. Each final lesson plan is validated by a team of pedagogical experts to ensure it meets our rigorous criteria for interdisciplinary linkage, graded difficulty, and task-driven objectives, while also aligning with contemporary teaching practices in the source context.

\noindent \textbf{Dialogue Generation Framework }
We construct a novel automated dialogue generation framework, centered on adversarial interaction between two LLM agents, supplemented by a rigorous expert review mechanism, aiming to generate high-quality, multi-turn Socratic dialogues at scale.

In this framework, we design two GPT-4o based LLM agents with distinct roles: the "teacher agent" acts as a Socratic mentor, equipped with designed lesson plans as instructional goals and background knowledge; the "student agent" is prompted to simulate six typical instructional challenges, reproducing common cognitive obstacles in real learning, such as failing to understand questions/explanations, lacking prior knowledge, showing little willingness to explore further despite preliminary answers, or having solid knowledge and higher-order thinking ability to enable in depth dialogues. Dialogues usually start with the student agent’s initial question, followed by at least five rounds of guided exploratory interaction, until the teacher agent deems the student has made a significant cognitive leap and generates a summary.

The most critical step in this process is the human expert review of all automatically generated dialogues. The educational experts review each round of dialogue to evaluate its instructional coherence, strategic validity, and factual accuracy. Dialogues that do not align with the principles of Socratic guidance or contain factual errors are excluded, ensuring the final dataset's high quality and pedagogical effectiveness.

\noindent \textbf{Annotation Schema}
To make complex dialogue interactions computationally tractable for model training and evaluation, we designed a 9-field multi-dimensional annotation schema, grounded in three complementary frameworks. First, from cognitive development, we use the Knowledge Integration Framework \cite{linn2006knowledge} to track epistemic processes like conflict identification and conceptual reorganization. Second, for dialogue dynamics, we extend the \ac{IRF} structure \cite{sinclair1975towards} to classify pedagogical moves such as probing assumptions and generating counterexamples. Third, based on Socratic pedagogy, we align with Bimantara's four-stage model \cite{bimantara2025cognitive} to map dialogue progression through conflict, comparison, restructuring, and synthesis phases.

The nine annotation fields systematically operationalize these theoretical constructs. 
\textbf{Speaker}: The role of the speaker.
\textbf{Utterance}: The original text of the turn.
\textbf{Teacher\_intent}: The teacher's pedagogical goal in the utterance (e.g., Concept Introduction, Comprehension Check, Reasoning Scaffolding, Knowledge Transfer Prompt, Summary).
\textbf{Teaching\_strategy}: The specific tactic used by the teacher (e.g., Follow-up Question, Hint, Analogy, Contextualization).
\textbf{Discipline}: The primary academic subject(s) involved in the utterance.
\textbf{discipline\_transfer}: A binary label indicating if a transfer of knowledge between disciplines is occurring.
\textbf{Student\_cognition\_state}: The inferred cognitive state of the student (e.g., Clear Understanding, Confusion, Misconception, High-level Synthesis).
\textbf{Teacher\_guidance\_level}: The level of the teacher's question (L1: Closed-ended, L2: Explanatory, L3: Open-ended/Inferential).
\textbf{Cognitive\_level}: The corresponding level of Bloom's Taxonomy for the utterance (e.g., Remember, Understand, Apply, Analyze, Evaluate, Create).

To ensure the transparency and reproducibility of our work, we provide comprehensive supplementary materials in the appendices. Appendix A contains detailed summary statistics of the SID dataset, a complete, fully annotated dialogue sample, the detailed annotation guidelines, and the results of our quality control measures. Appendix D further details our dialogue generation framework, including the core prompts used for the teacher and student agents.

We implement a rigorous, multi-stage quality control process to maintain the high quality of our dataset. For annotation accuracy, we use a pipeline that combines LLM-based pre-annotation with human expert verification. A randomly selected 20 percent of the data is spot-checked by our team of pedagogical experts. To ensure content safety, we also apply automated toxicity detection to all generated dialogues to remove any inappropriate content.

\subsection{Evaluation Framework}
\noindent To systematically assess the performance of LLMs in guided pedagogical dialogues, we design a two-tier evaluation framework for a comprehensive assessment of the quality of interaction. This framework combines quantifiable pedagogical behaviors (via objective metrics) with holistic dialogue quality (via subjective Rubrics), allowing us to assess the strategic quality of teacher guidance, the depth of student cognitive growth, and the effectiveness of interdisciplinary reasoning.

\noindent \textbf{Objective Behavioral Indicators }
Based on our structured annotation framework, we define a set of seven automatically computable metrics aimed at capturing teaching effectiveness from multiple dimensions. The design of these metrics is inspired by the Socratic teaching theory~\cite{murray1999socratic}, the Conceptual Change Model~\cite{chi1993conceptual}, and Bloom’s taxonomy of cognitive development~\cite{bloom1956taxonomy}, and draws upon evaluation frameworks proposed in prior studies such as SocraticLM~\cite{liu2024socraticlm} and M3KE~\cite{liu2023m3ke}. Specifically, the metrics include: 
(1) \ac{SD}, the frequency of teaching strategies; (2) \ac{SV}, the number of unique strategies used; (3) \ac{SC}, whether the dialogue covers the core pedagogical intents; (4) \ac{L3 GR}, the proportion of high-level (L3) questions; (5) \ac{3C}, the number of instances where a student transitions from confusion or misconception to understanding; (6) \ac{BP}, the average increase in the Bloom's Taxonomy level of the student's utterances; and (7) \ac{IKT}, the count of explicit instances where interdisciplinary knowledge transfer occurs in the dialogue.

Our weighting scheme is designed to establish a student-centric evaluation paradigm that balances pedagogical process with learning outcomes, distinguishing our approach from prior work. Benchmarks like SocraticLM, for instance, primarily emphasize the conformance of teacher guidance strategies, while M3KE emphasizes the structure of multi-step reasoning paths. Our philosophy posits that a successful pedagogical dialogue requires not only an exemplary guidance process from the teacher but must also result in observable cognitive growth in the student.

Therefore, we distribute the weights across three core dimensions. the Teacher's Pedagogical Process (50\%) as a prerequisite for success. Student's Cognitive Growth (30\%) as the ultimate goal of the instruction, with Cognitive Correction receiving the highest individual weight (20\%) as it is the essence of Socratic guidance. Interdisciplinary Synthesis (15\%) to reflect the core challenge of our benchmark. This philosophy is implemented in the following composite score formula:
\begin{eqnarray}
& \text{TotalScore} = \sum_{i=1}^{7} w_{i} \cdot \text{metric}_{i} 
\end{eqnarray}

The specific weights ($w_{i}$) are detailed in our Appendix B. This balanced scheme provides a more holistic and pedagogically-grounded assessment of an LLM's teaching capabilities.

\noindent \textbf{Subjective Quality Rubrics }
To complement the objective metrics by capturing holistic and strategic qualities, we design five Rubric-based indicators and employ an "LLM-as-a-Judge" approach for automated evaluation. We utilize DeepSeek-V3  as the judge, providing it with the full dialogue transcript and detailed scoring criteria for each Rubric (on a 5-point Likert scale) via a carefully designed prompt. To validate the reliability of our judge model, we invite pedagogical experts to conduct a parallel scoring on 200 dialogue samples. The experimental results show a high Pearson correlation between DeepSeek-V3 's scores and the average scores from human experts, confirming the validity of this automated evaluation method.

These indicators include: (1) Interdisciplinary Scientific Reasoning Grading (X-SRG), (2) Multi-turn Reasoning Coherence (M-RCC), (3) \ac{X-MSR}, (4) \ac{IRA}, and (5) \ac{TCF}. These Rubrics are designed to evaluate higher-order capabilities that are difficult to capture with automated metrics alone.

To ensure transparency and reproducibility, Appendix B details our evaluation framework, including objective metric weightings, subjective rubric criteria, and core prompts for "LLM-as-a-Judge" and semi-automated annotation.

%% file: 04.Experimental.tex
\section{Experiments \label{sec:experiment}}
\subsection{Experimental Setup}
\subsubsection{Models}
\noindent We evaluate six \ac{LLMs} on our benchmark dataset, including three general-purpose models and three education-oriented models. Detailed hyperparameter settings for all baseline models are provided in Appendix C.

\textbf{GPT-4o:}  
Fully multimodal, GPT-4o efficiently processes text, audio, and images~\cite{hurst2024gpt}, with improved fairness for non-English languages and strong potential in medical/scientific applications.
\textbf{QwQ-32B:}  
A 32B-parameter model fine-tuned via reinforcement learning for enhanced reasoning, supporting tool invocation and dynamic reasoning adjustment~\cite{team2025qwq}.
\textbf{Qwen-2.5-14B-Instruct:}  
Instruction-tuned to better follow precise directives, boosting performance in structured text generation and targeted questions~\cite{qwen2.5}.
\textbf{SocraticLM:}  
Based on ChatGLM3-6B~\cite{du2022glm}, this educational LLM uses Socratic open-ended questions to foster critical thinking~\cite{liu2024socraticlm}.
\textbf{EduChat-R1:}  
A hybrid-reasoning educational LLM employing "Educational Chain-of-Thinking" to simulate teacher reasoning, shifting from knowledge delivery to guided reflection.
\textbf{InnoSpark:}  
Built on Qwen2.5-72B~\cite{qwen2.5}, it integrates Chinese educational values to promote creativity and personalized learning, facilitating collaboration among teachers, students, and LLM\footnote{\url{https://huggingface.co/sii-research/InnoSpark-72B-0710}}.

\subsubsection{Evaluation}
This study adopts a dual evaluation framework comprising both subjective and objective measures to comprehensively and rigorously assess the true capabilities of LLMs in Socratic interdisciplinary guidance tasks. Furthermore, the dataset used in this study is built within a Chinese language context.

\begin{figure}
    \centering
    \small
    \includegraphics[width=6cm]{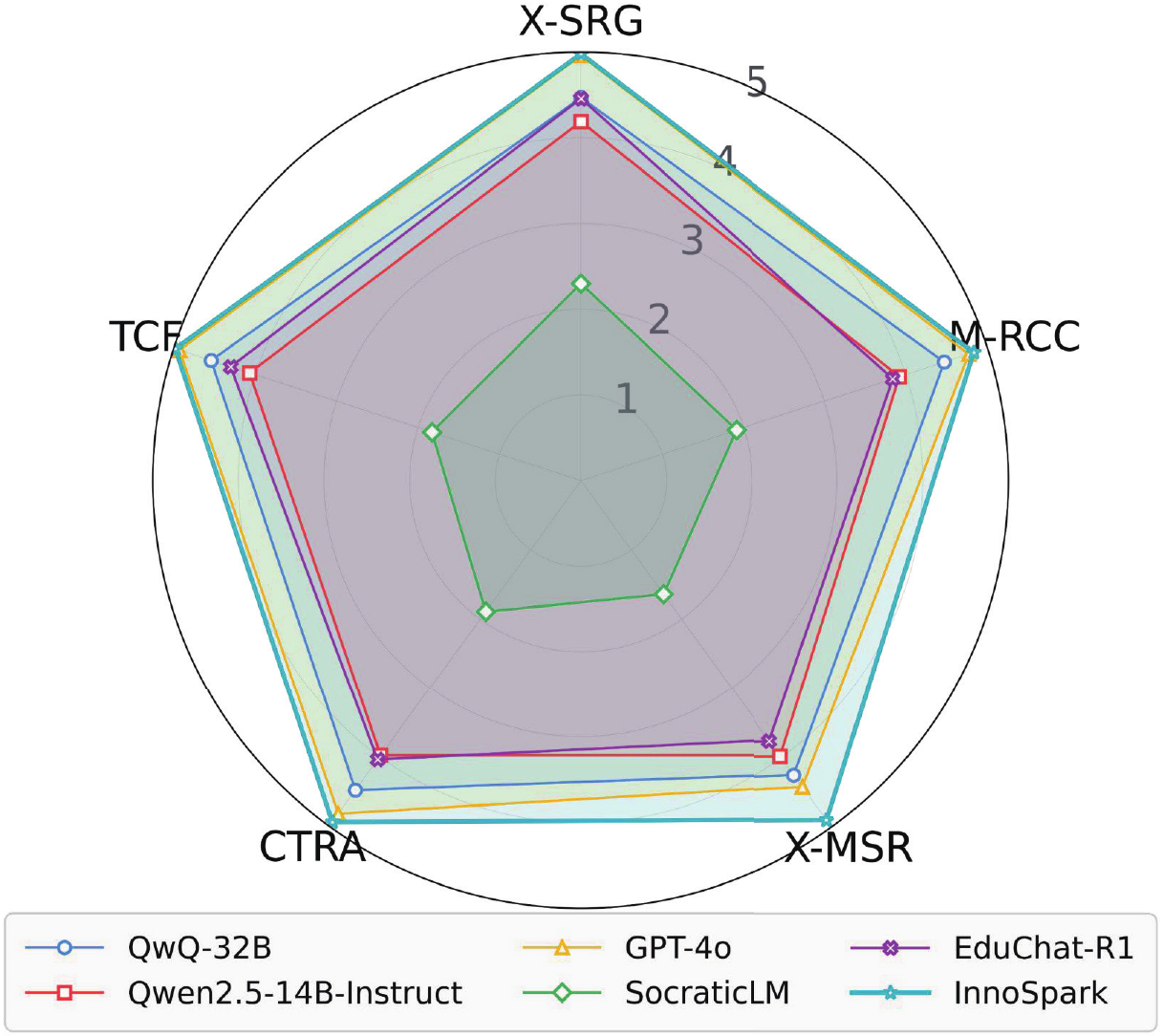}
    \caption{Subjective Evaluation of the SID Benchmark.}
    \label{fig:subjective}
\end{figure}
In the subjective evaluation, we employ DeepSeek-V3 as an automated judge model, rate the teacher model’s performance on a 1 to 5 scale across five predefined core metrics, and require it to output scores and rationales in strict JSON format.

DeepSeek‑V3 is chosen as the judge model primarily for its relative independence, which helps reduce the risk of self‑bias, ensuring objective and automation‑friendly evaluation results.

Objective evaluation provides quantifiable, reproducible, and scalable metrics via a multi-dimensional framework encompassing pedagogical strategy utilization (\ac{SD}, \ac{SV}), interdisciplinary knowledge integration and cognitive advancement (\ac{IKT}, \ac{BP}, \ac{L3 GR}), and dialogue structuring with cognitive error correction (\ac{SC}, \ac{3C}), quantifying LLM teacher efficacy.

To enable large-scale automated annotation, we use Qwen3-32B~\cite{yang2025qwen3} for structured dialogue labeling, as it offers strong task adaptability, compatibility with Chinese educational contexts, and a balance of efficiency and accuracy for downstream analysis.

\subsection{Main Results}
\noindent This section presents and compares \ac{LLMs}' performance under different evaluation dimensions.

The subjective evaluation of \ac{LLMs} on the \ac{SID} benchmark, as detailed in Table \ref{tab:subjective} and Fig \ref{fig:subjective}, reveals a clear hierarchy in their ability to perform Socratic interdisciplinary guidance. InnoSpark consistently excels, leading in most subjective metrics and showcasing strong Socratic guidance and interdisciplinary reasoning.

Next, we present the quantitative results of the objective evaluation(Table \ref{tab:objective}). 
QwQ-32B ranks highest overall, excelling in strategy variety and cognitive advancement. GPT-4o follows closely, strong in knowledge transfer, high-level questioning, and misconception correction.
\begin{figure*}[t]
    \centering
    \includegraphics[width=1.0\textwidth]{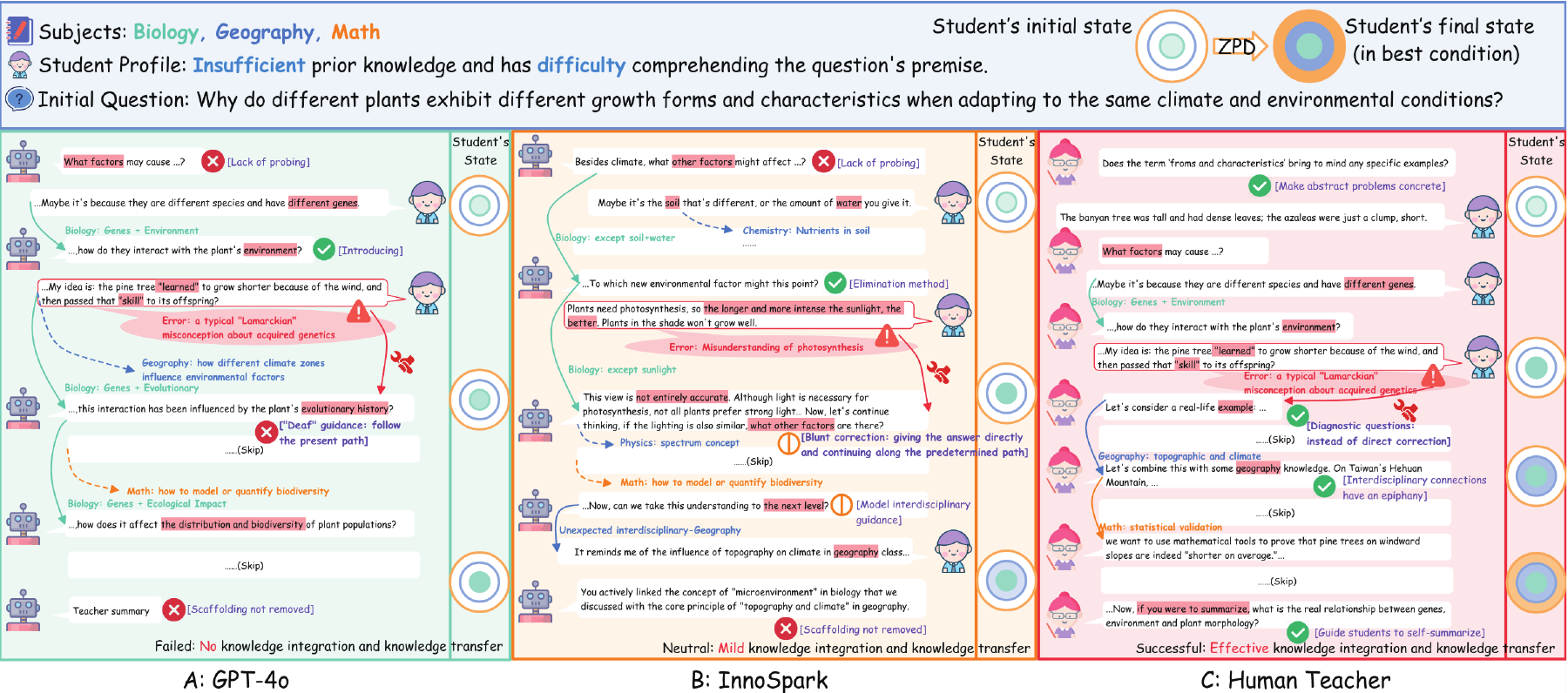}
    \caption{\textbf{A comparative case study of guidance strategies from LLM Teachers and a Human Expert.} Solid arrows denote actual dialogue pathways, tracing the progression of topics and the primary academic discipline of each turn.
Dotted arrows denote potential, ideal pedagogical pathways that were not taken, such as missed opportunities for interdisciplinary connections.
Text in purple annotates the specific pedagogical strategy (e.g., Diagnostic Questioning) corresponding to a teacher's utterance.
LLM tutors (Panels A and B) demonstrate rigid guiding strategies: GPT-4o (Panel A) ignores students' critical errors in order to follow its linear "golden path"; InnoSpark (Panel B) rigidly chooses to directly correct students' mistakes. In sharp contrast, the human expert (Panel C) shows a dynamic, diagnostic approach, successfully turning students' misconceptions into teaching opportunities and establishing organic interdisciplinary connections.}
    \label{fig:casestudy}
\end{figure*}

\begin{table}[t]
  \centering
  \scriptsize
  \setlength{\tabcolsep}{2pt}
  \renewcommand{\arraystretch}{1.2}
  \begin{tabularx}{\columnwidth}{l *{6}{>{\centering\arraybackslash}X}}
    \toprule
    Metric & \textbf{GPT-4o} & \textbf{QwQ} & \textbf{Qwen} & \textbf{SocLM} & \textbf{EduChat} & \textbf{InnoSpark} \\
    \midrule
    \ac{X-SRG}       & \underline{4.96} & 4.47 & 4.19 & 2.30 & 4.46 & \textbf{4.98} \\
    \ac{M-RCC}       & \underline{4.77} & 4.47 & 3.92 & 1.91 & 3.84 & \textbf{4.83} \\
    \ac{X-MSR}       & \underline{4.42} & 4.25 & 3.98 & 1.64 & 3.76 & \textbf{4.90} \\
    \ac{IRA}         & \underline{4.82} & 4.48 & 3.98 & 1.89 & 4.04 & \textbf{4.94} \\
    \ac{TCF}         & \underline{4.93} & 4.54 & 4.07 & 1.83 & 4.30 & \textbf{4.98} \\
    AvgScore         & \underline{4.78} & 4.44 & 4.03 & 1.91 & 4.08 & \textbf{4.93} \\
    \bottomrule
  \end{tabularx}
  \caption{\textbf{Subjective Evaluation of the SID Benchmark.} "QwQ", "Qwen", "SocLM" and "EduChat" stand for QwQ-32B, Qwen2.5-14B-Instruct, SocraticLM and EduChat-R1. 
  The max score of all these metrics is 5.0.}
  \label{tab:subjective}
\end{table}

\begin{table}[ht]
  \footnotesize
  \centering
  \scriptsize
  \setlength{\tabcolsep}{2pt}
  \renewcommand{\arraystretch}{1.2}
  \begin{tabularx}{\columnwidth}{l *{6}{>{\centering\arraybackslash}X}}
    \toprule
    Metric & \textbf{GPT-4o} & \textbf{QwQ} & \textbf{Qwen} & \textbf{SocLM} & \textbf{EduChat} & \textbf{InnoSpark} \\
    \midrule
    SD       & \textbf{99.73\%} & 98.43\% & 98.84\% & 97.85\% & 99.67\% & 99.59\% \\
    SV       & 55.76\% & \textbf{81.75\%} & 43.23\% & 39.73\% & 57.64\% & \underline{59.99\%} \\
    IKT      & \textbf{40.24\%} & 28.09\% & 24.04\% & \underline{32.51\%} & 18.36\% & 21.94\% \\
    BP       & 32.37\% & \textbf{55.17\%} & 36.10\% & 22.17\% & 28.46\% & \underline{38.66\%} \\
    SC       & 74.34\% & 73.82\% & 52.81\% & 48.71\% & \textbf{77.87\%} & \underline{75.99\%} \\
    L3~GR    & \textbf{81.47\%} & 65.83\% & 43.67\% & 44.31\% & 52.16\% & \underline{77.64\%} \\
    3C       & \underline{99.90\%} & 98.54\% & \textbf{99.93\%} & 96.11\% & 99.55\% & 99.79\% \\
    TotalScore & \underline{0.7070} & \textbf{0.7279} & 0.6045 & 0.5781 & 0.6455 & 0.6915 \\
    \bottomrule
  \end{tabularx}
  \caption{\textbf{Objective evaluation of the SID benchmark.} "QwQ", "Qwen", "SocLM" and "EduChat" stand for QwQ-32B, Qwen2.5-14B-Instruct, SocraticLM and EduChat-R1.}
  \label{tab:objective}
\end{table}

\subsection{Analysis}
\subsubsection{Subjective Evaluation}

\noindent Based on the subjective evaluation results presented in Table\ref{tab:subjective}, we conduct an in‑depth analysis of the performance of various \ac{LLMs} on the Socratic interdisciplinary guided dialogue task.

LLMs face inherent challenges in achieving genuine interdisciplinary integration and facilitating deep knowledge transfer: high-performance models deliver coherent interactions yet remain superficial in interdisciplinary connections, moderate-performance models exhibit inconsistent performance across interdisciplinary tasks, and low-capability models lack the advanced reasoning capacity to support such integration.

(1) InnoSpark leads across all five evaluation metrics, excelling particularly in multi-turn guidance, interdisciplinary reasoning, and dialogue fluency. GPT-4o follows closely, matching InnoSpark on \ac{X-SRG} and \ac{TCF}, a reflection of its strong general reasoning and coherent dialogue abilities.
(2) QwQ-32B performs strongly on \ac{M-RCC}, X-MSR, \ac{IRA}, and \ac{TCF}, with M-RCC rivaling GPT-4o and InnoSpark. Qwen2.5-14B-Instruct shows stable performance and some Socratic guidance capability but still trails overall.
(3)SocraticLM performs poorly across all metrics, especially in reasoning and interdisciplinary integration, making it ill-suited for complex teaching—likely due to its focus on math instruction in English contexts. EduChat-R1 has moderate capabilities, with room to refine its scaffolding of complex learning pathways.
\subsubsection{Objective Evaluation}
\noindent Based on the objective evaluation results in Table \ref{tab:objective}, we analyze the models’ performance in the interdisciplinary Socratic teaching task.

LLMs generally perform poorly in interdisciplinary knowledge integration and deep transfer. High-performance models struggle with holistic integration, moderate-performance models lack higher-order transfer abilities, and low-capability models miss basic integration mechanisms.
(1) QwQ-32B leads overall due to its strong pedagogical strategies and cognitive scaffolding. GPT-4o closely follows, excelling in interdisciplinary reasoning and high-level questioning. Notably, GPT-4o achieves an almost 0 \ac{3C}, highlighting its near-perfect ability to prevent or promptly rectify student misconceptions.
(2) InnoSpark and EduChat-R1 both demonstrate robust teaching capabilities, each with distinct strengths. InnoSpark excels in high-order guidance metrics like \ac{SV} and \ac{BP}, reflecting its comprehensive pedagogical approach. EduChat-R1 achieves the highest score in \ac{SC} and shows superior \ac{3C}, highlighting its advantage in building clear teaching flows and addressing student cognitive biases.
(3) In contrast, SocraticLM and Qwen2.5-14B-Instruct exhibit significant deficiencies in core scaffolding capabilities. Despite a high \ac{SD}, SocraticLM records the lowest \ac{SV} and performs poorly in metrics reflecting higher-order thinking cultivation, such as \ac{BP} and \ac{SC}. Despite a high \ac{SD}, Qwen2.5-14B-Instruct shows substantial room for improvement in \ac{SC} and \ac{L3 GR}. While it achieves the highest \ac{3C}, its low TotalScore indicates that its overall scaffolding effectiveness needs strengthening.
\subsection{Case Study}
\noindent To provide a nuanced, qualitative analysis of current model capabilities, we conducted a comparative case study. We analyzed the guided dialogues of a general-purpose LLM (GPT-4o) and a specialized educational LLM (InnoSpark) on the same interdisciplinary STEM pedagogical task, and contrasted their performance with the guidance strategies of an ideal human expert. Fig. \ref{fig:casestudy} visually summarizes these three distinct dialogue pathways. The more cases can be found in Appendix E.

\noindent \textbf{Model Capability Profiles}
\begin{itemize}
\item \textbf{General-Purpose Large Language Model (GPT-4o): }
GPT-4o demonstrates a logically clear yet highly linear "golden path" guiding approach. Its advantage lies in its ability to construct a rigorously structured cognitive ladder, guiding the dialogue to gradually delve from "genes" into "environment" and "evolution" through a series of coherent follow-up questions. However, it has a profound flaw in its "deaf" guidance: when the student reveals a typical Lamarckian misconception, the LLMs completely ignore this crucial teaching opportunity and continues to execute its pre-set script. Although this externally results in eventually guiding the student to obtain the correct answer to the initial question, it fails to correct the errors made in the process. Moreover, its guiding path is confined within biology, missing the potential opportunities for organic connections with geography, mathematics, and other disciplines. As a result, the student ultimately fails to step beyond their zone of proximal development, only acquiring biological knowledge without effectively gaining the ability to integrate and transfer knowledge.

\item \textbf{Specialized Educational Large Language Model (InnoSpark): }
The specialized educational model InnoSpark, on the other hand, demonstrates a style with stronger awareness of teaching strategies, but it also involves trade-offs. This model successfully employs the structured scaffold of "elimination method," breaking down complex problems into a series of manageable small steps (such as soil, light, wind, etc.). More importantly, it successfully identifies and intervenes in students' misconception that "the stronger the light, the better." However, its way of correcting errors is non-Socratic, adopting a blunt strategy of "direct negation + knowledge inculcation," which interrupts the students' inquiry process. In terms of interdisciplinary guidance, like GPT-4o, its own guiding path is completely confined to a single discipline; the most critical connection with geography in the dialogue is spontaneously established by the student through rigidly patterned questions, and the LLM only acts as a passive identifier rather than an active facilitator in this process. Such an approach prevents students from effectively stepping beyond their zone of proximal development, resulting in only a slight improvement in their abilities of knowledge integration and transfer.
\end{itemize}

\noindent \textbf{Comparative Analysis and The Gap with Human Teachers}
This comparison reveals different failure modes of current LLM tutors: GPT-4o completely ignores the student's error, while InnoSpark identifies the error but handles it inappropriately. Both demonstrate deficiencies in proactively guiding interdisciplinary connections.

When compared to expert human teachers, all LLMs pale in comparison. Human teachers embody a dynamic adaptability that LLMs currently cannot match. An expert teacher can not only diagnose and utilize students' misconceptions as teaching opportunities (for example, through carefully designed counterexamples) but also proactively and organically connect topics with different disciplines such as geography and mathematics. Most crucially, human teachers can flexibly switch between different guiding strategies based on students' real-time responses, and ultimately remove the scaffolds to guide students to construct and summarize knowledge on their own. Eventually, this enables students to step beyond their zone of proximal development and acquire the abilities of knowledge integration and transfer.

This case study demonstrates that while LLMs have become considerably capable of replicating the "form" of structured guidance, with general-purpose models excelling at logical deduction and specialized models beginning to attempt error identification and strategy decomposition, a fundamental gap remains in capturing the "essence" of teaching. This essence lies in the dynamic adaptability to handle pedagogical contingencies, the interdisciplinary creativity to foster knowledge integration, and the deep diagnosis of students' cognitive processes. This strongly validates the need for developing a benchmark like SID, which is capable of measuring these deep, multi-dimensional pedagogical capabilities, to drive the field of LLM in Education forward.

%% file: 05.Conclusions.tex
\section{Conclusion \label{sec:conclusion}}
\noindent In this work, we introduce \textbf{S}ocratic \textbf{I}nterdisciplinary \textbf{D}ialogues (SID), a novel benchmark and evaluation framework designed to rigorously assess LLMs' capabilities in Socratic interdisciplinary dialogue for STEM education. SID includes over 10,000 dialogue turns across 48 complex STEM projects, supported by a multi-dimensional annotation schema and comprehensive metrics. Our evaluations reveal that despite advancements, current LLMs—whether general-purpose or education-oriented—struggle with dynamic pedagogical adaptation, deep interdisciplinary integration, and effective scaffolding of students' knowledge transfer. This highlights the critical need for further research to develop truly pedagogically-aware LLMs. SID serves as a foundational tool to drive progress in this direction, facilitating the development of AI tutors that can effectively foster students' ability to integrate and transfer knowledge.